\documentclass[11pt]{article}

\usepackage[letterpaper,margin=1in]{geometry}
\usepackage[T1]{fontenc}
\usepackage{times}
\usepackage{microtype}
\usepackage{amsmath,amsfonts,bm}

\def\eqref#1{equation~\ref{#1}}
\def\Eqref#1{Equation~\ref{#1}}

\def\1{\bm{1}}

\DeclareMathAlphabet{\mathsfit}{\encodingdefault}{\sfdefault}{m}{sl}
\SetMathAlphabet{\mathsfit}{bold}{\encodingdefault}{\sfdefault}{bx}{n}

\usepackage[round]{natbib}
\usepackage{graphicx}
\usepackage{xcolor}
\usepackage{capt-of}
\usepackage{longtable}
\usepackage{array}
\usepackage{booktabs}
\usepackage{tabularx}
\usepackage{multirow}
\usepackage{colortbl}
\usepackage{pgf}
\usepackage{xspace}
\usepackage{url}
\usepackage[font=small,labelfont=bf,skip=6pt]{caption}
\usepackage[colorlinks=true,linkcolor=blue!60!black,citecolor=blue!60!black,urlcolor=blue!60!black]{hyperref}

\usepackage{titlesec}
\titlespacing*{\section}{0pt}{2.2ex plus 1ex minus .2ex}{1.2ex plus .2ex}
\titlespacing*{\subsection}{0pt}{1.8ex plus 1ex minus .2ex}{0.8ex plus .2ex}
\titlespacing*{\paragraph}{0pt}{1.2ex plus .5ex minus .2ex}{1em}

\newcommand{\score}[1]{%
  \pgfmathtruncatemacro{\shadeval}{4 + 0.30*(#1)}%
  \edef\tempcolor{red!\shadeval}%
  \expandafter\cellcolor\expandafter{\tempcolor}#1\%
}

\newcommand{\scoreb}[1]{%
  \pgfmathtruncatemacro{\shadeval}{4 + 0.30*(#1)}%
  \edef\tempcolor{red!\shadeval}%
  \expandafter\cellcolor\expandafter{\tempcolor}\textbf{#1\%}%
}

\definecolor{reusedc}{HTML}{2F6DB5}
\definecolor{newc}{HTML}{E07B00}
\newcommand{\method}{ECoMEM\xspace}

\makeatletter
\renewcommand{\@maketitle}{%
  \newpage
  \null
  \vspace{0.5em}%
    \begin{center}
    {\LARGE\bfseries \@title \par}%
    \vspace{1.4em}%
    {\large \@author \par}%
    \vspace{0.6em}%
    {\large \url{https://ecomem.github.io/} \par}%
  \end{center}
  \vspace{1.2em}%
  \begin{center}
    \includegraphics[width=\linewidth]{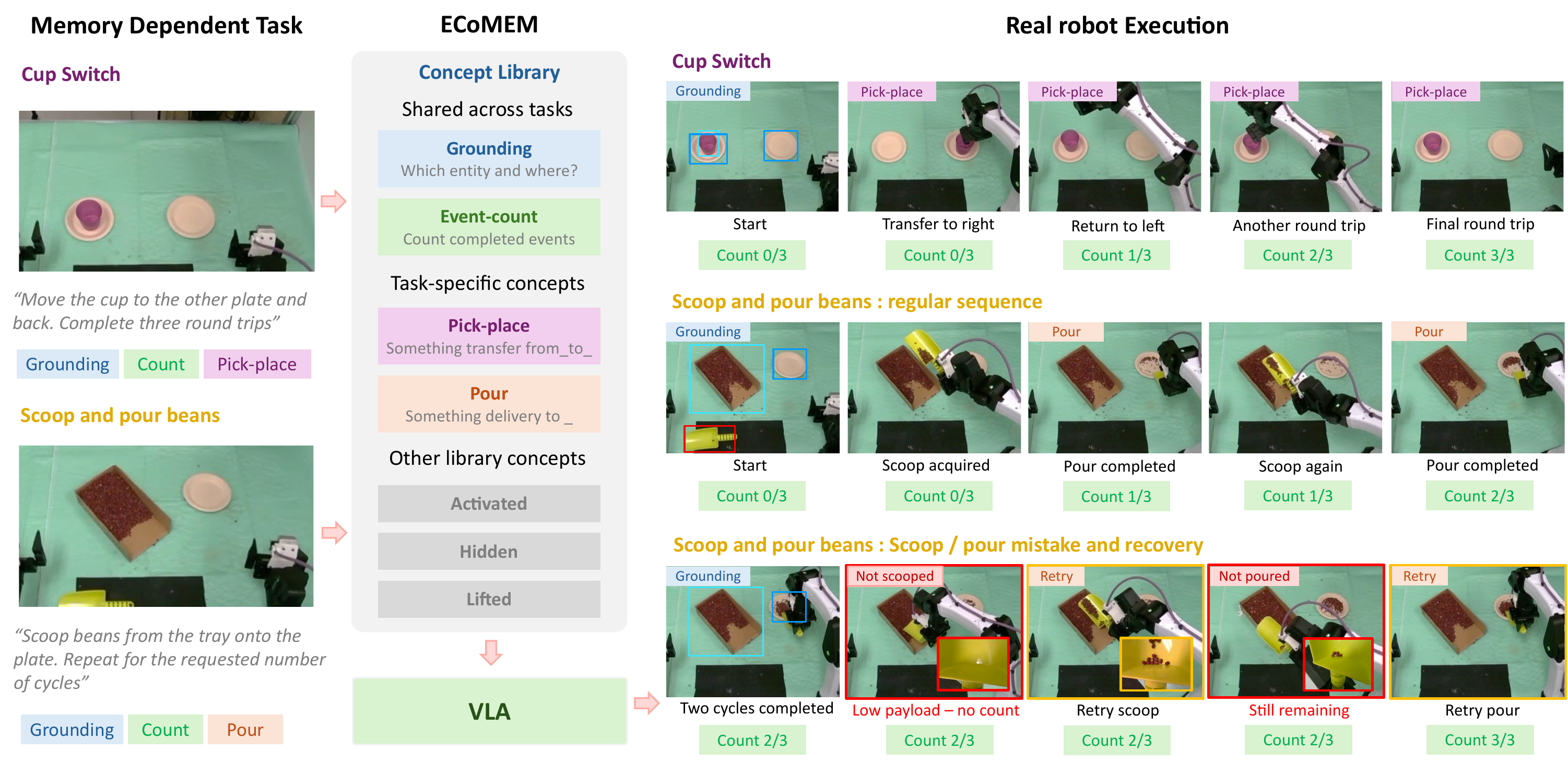}
    \captionof{figure}{\textbf{Explicit Concept Memory (\method)} represents task-relevant history with compositional concept memory.
\textbf{Left}: memory-dependent tasks require information that may no longer be available in the current observation.
\textbf{Middle}: shared concepts for grounding and progress compose with task-specific event concepts.
\textbf{Right}: the resulting memory conditions the VLA to track progress and recover from unsuccessful events on a real robot.}
    \label{fig:teaser}
  \end{center}
  \vspace{0.5em}%
}
\makeatother

\title{\method: Explicit Concept Memory for\\Memory-Dependent Robot Control}

\author{%
  Yize Liu \quad Ke Wang \quad Mac Schwager \quad Yiqing Xu$^{\dagger}$ \quad Jiajun Wu$^{\dagger}$ \\[0.4em]
  {\normalsize Stanford University}
  \\[0.2em] {\normalsize\texttt{\{yizeliu,yiqingx\}}@stanford.edu}
  \\[0.2em] {\small $^{\dagger}$Equal advising}
 }

\date{}

\begin{document}

\maketitle

\begin{abstract}
A robot may lose sight of an object it must later retrieve, need to recall
what a person demonstrated earlier, or track which steps of a task it has
already completed. Current vision--language--action (VLA) policies often
fail once the information needed for action disappears from the current
observation, making memory critical for long-horizon robot behavior.
Existing approaches typically provide longer histories or learn implicit
memory from observation--action trajectories. But action supervision tells
a policy \emph{how to act}, not \emph{what to remember}: it does not specify
which past facts should persist or how they should change as new evidence
arrives. We therefore separate \emph{maintaining an evidence-grounded account of
the past} from \emph{learning how to act on it}. This insight motivates
\textbf{Explicit Concept Memory (\method)}, which represents task-relevant
history with a reusable library of grounded concepts. An evidence-based
Writer selects and updates these records, while a learned Reader turns
them into memory tokens that directly condition the VLA. Across 16 RoboMME tasks, \method leads the evaluated robot policies on
15 tasks. On two new real-robot tasks, the same memory library either
transfers directly or requires only one new concept, achieving $86.1\%$
success versus $8.6\%$ for a no-memory VLA. These results show that
explicit concepts provide a reusable and extensible memory interface for
robot control. \href{https://ecomem.github.io/}{Project website}.
\end{abstract}

\section{Introduction}
\label{sec:intro}

A robot that acts over extended periods cannot treat every observation
as a fresh start. It may need to remember where an object was last seen
after it becomes occluded, which object a person interacted with earlier,
what parts of a task have already been completed, or the order in which
previous events occurred. In each case, the information needed for the
next action is no longer available in the current observation. Yet the
relevant past is often much smaller than the full interaction history:
what matters is a compact set of task-relevant facts about entities,
states, events, and progress.

Vision--language--action (VLA) policies increasingly incorporate history
through extended context, recurrent representations, learned memory
banks, and test-time adaptation~\citep{ptp,hamlet,memoryvla,ttt}.
These mechanisms can preserve rich perceptual history, but they leave an
important question implicit: \emph{what from that history should become
memory?} Action supervision specifies what the robot should do, but not
which past facts should be retained, when a belief should be revised, or
when two observations correspond to the same event. When memory is
learned only through the control objective, the policy may instead rely
on shortcuts in current observations, previous actions, or
trajectory-specific correlations~\citep{causalconfusion,copycat,
proprioshortcut}. Successful action prediction therefore does not by
itself establish that the task-relevant history has been represented
reliably.

Making memory explicit helps, but does not remove the problem of memory
construction. Keyframe memories preserve visual evidence
~\citep{eventvla,fang2025sam2act}, leaving the policy to recover the
relevant facts from stored images. VLM-based approaches summarize
histories~\citep{sridhar2026memer,remembr}, but place entity tracking,
state estimation, and temporal interpretation inside long-video
reasoning, where spatial and temporal errors remain challenging
~\citep{beyondsemantics,egoillusion}. Textual world states
~\citep{statler,liu2023reflect} and scene graphs
~\citep{gu2024conceptgraphs,sayplan,dynamem} provide more structured
representations, but a robotic memory must additionally decide how
observations become persistent facts: what should survive occlusion,
what should be revised, what constitutes a completed event, and how
repeated evidence should be reconciled.

Our key idea is to separate \emph{maintaining an evidence-grounded
account of the past} from \emph{learning how to act on it}. Rather than
asking the policy to infer memory implicitly from history, we represent
task-relevant history as explicit facts about entities, states,
completed events, progress, and temporal order. These facts are updated
only when supported by visual and robot-state evidence, while the policy
learns how they should influence action. This separation also keeps task
requirements distinct from observations: the instruction specifies what
should be achieved, while memory records what has actually happened.

We instantiate this principle in \textbf{Explicit Concept Memory
(\method)}, an explicit memory channel for VLA policies
(Figure~\ref{fig:org}). First, a shared concept library provides reusable
memory primitives spanning entity and spatial grounding, state and
relations, events and progress, and temporal and procedural structure.
Second, an evidence-based \emph{Writer} selects the concepts required by
the instruction, grounds them to task entities, and maintains their
records over time, while a learned \emph{Reader} maps the resulting
structured memory into latent tokens. Third, these memory tokens directly
condition the VLA alongside visual and language inputs. Recognition and
memory writing remain fixed during policy training, so evidence determines
\emph{what is remembered}, while action supervision learns \emph{how that
memory should be used}.

Across the 16 memory-dependent tasks of RoboMME~\citep{dai2026robomme},
\method achieves the highest average success among the evaluated robot
policies and leads on 15 of 16 tasks. Controlled experiments show that
spatial grounding, event and progress information, temporal structure,
task-conditioned selection, and language grounding each contribute to
effective memory construction. More importantly, the same abstraction
transfers to new physical-robot tasks: \textsc{CupSwitch} reuses the
existing library without adding a new recognizer, while
\textsc{ScoopPour} requires only one new event concept composed with the
existing memory machinery. Across shared real-robot conditions, \method
succeeds in $68/79$ trials ($86.1\%$), compared with $5/58$ ($8.6\%$)
for the no-memory $\pi_{0.5}$ baseline. These results suggest that
explicit, compositional concepts can serve not merely as task-specific
memory annotations, but as a reusable and extensible interface between
a robot's accumulated experience and learned control.

\section{Related Work}
\label{sec:related}

\subsection{Implicit memory in robot policies}
History-aware robot policies typically learn a latent representation of
past observations and actions. Long-context policies directly attend to
extended trajectories~\citep{ptp,hamlet}; recurrent architectures propagate
learned state across time~\citep{rememvla}; memory-bank methods compress and
retrieve past perceptual features~\citep{memoryvla}; and test-time adaptation
stores history in model weights~\citep{ttt}. Predictive world models similarly
maintain latent state for control~\citep{daydreamer,r2i}. These approaches can
capture rich temporal dependencies without prescribing what the representation
must contain. However, what is retained, forgotten, or updated is learned
largely through downstream objectives and remains difficult to inspect
independently of the policy. ECoMEM instead externalizes a sparse set of
task-relevant facts and fixes their evidence-based updates, while learning only
how the policy should use them. Explicit and latent memory are therefore
complementary rather than mutually exclusive.

\subsection{Explicit memory in robot policies}

A second line of work stores selected parts of experience explicitly.
Keyframe and retrieval methods retain event-triggered images or trajectory
segments~\citep{eventvla,fang2025sam2act,strap,vid2robot}; MemER retrieves
task-relevant visual experiences and uses a high-level VLM to condition a
low-level policy~\citep{sridhar2026memer}. Other approaches summarize experience into language
~\citep{remembr,liu2023reflect}, maintain textual world states
~\citep{statler}, or organize persistent information in scene graphs
~\citep{gu2024conceptgraphs,sayplan,dynamem}. These representations improve
access to past information, but visual memories still require the controller
to recover facts from stored observations, while language and graph memories
are commonly consumed by high-level reasoning or planning modules. ECoMEM
instead maintains typed, evidence-linked records of states, events, progress,
and temporal structure, and feeds them directly to a low-level VLA through a
learned Reader.

\subsection{Neuro-symbolic methods in robotics}

Neuro-symbolic robotics combines learned perception and control with
explicit, compositional abstractions. Knowledge representations such as
KnowRob organize persistent robot knowledge~\citep{tenorth2009knowrob};
neuro-symbolic concept learning and structured world models ground
symbolic concepts in perception and compose them for reasoning
~\citep{mao2019nscl,pdsketch}; and learned or invented predicates provide compact
abstractions for task planning
~\citep{visualpredicator,pix2pred,predicateinvention}. Related work also
uses symbolic concepts to structure learned generative models for
manipulation and geometric reasoning~\citep{setitup,stackitup}.
Recent systems further use vision--language models to construct symbolic
planning abstractions from demonstrations~\citep{buildonpriors}, while
knowledge graphs provide structured memory for language-model agents
~\citep{arigraph}. ECoMEM follows the same principle of grounding reusable symbolic concepts
in learned representations, but uses them for a different role: as the
interface between perceptual history and a learned control policy.

\section{Preliminaries}
\label{sec:prelim}

\begin{figure*}[!t]
  \centering
  \includegraphics[width=\linewidth]{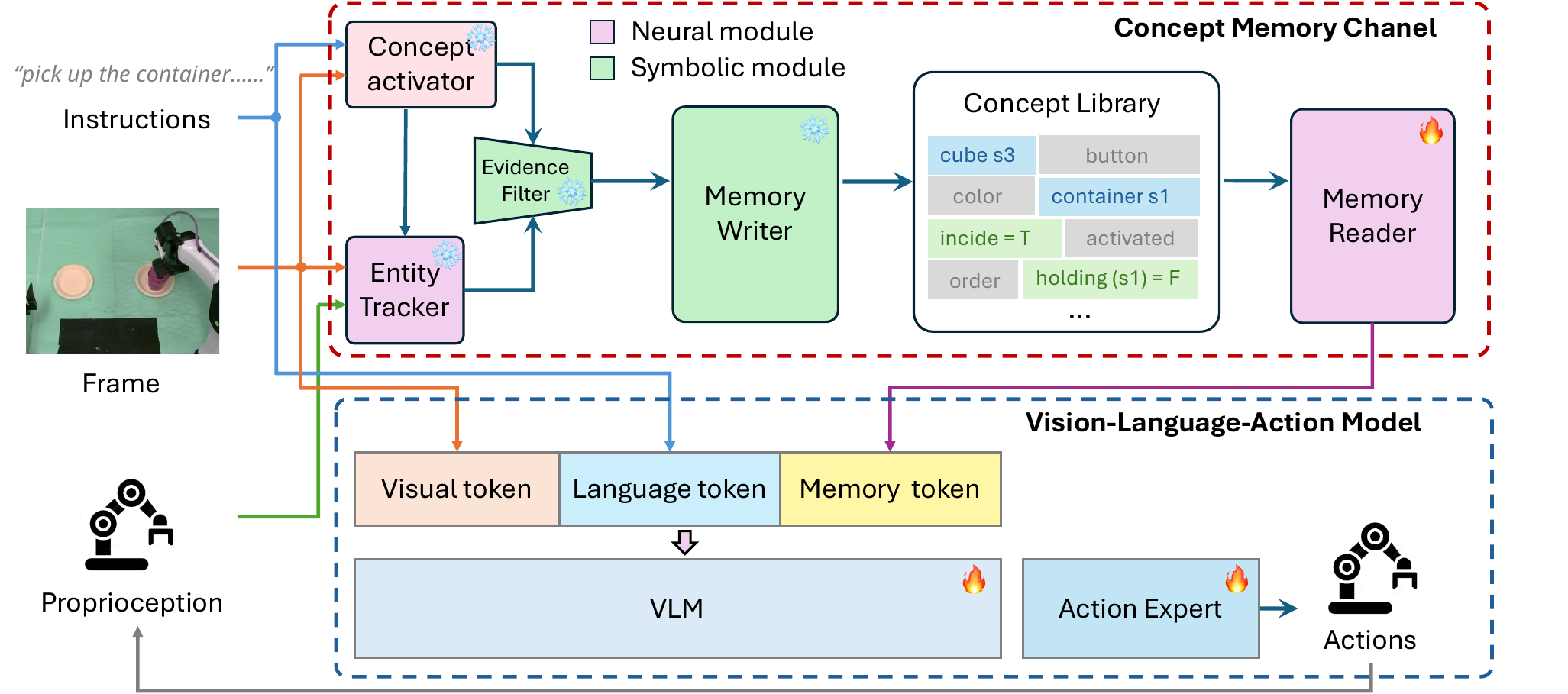}
  \caption{\textbf{\method separates memory construction from memory use.}
The Writer selects task-relevant concepts, recognizes them from visual and robot-state evidence, and maintains structured records.
The learned Reader encodes these records as memory tokens that condition the VLA alongside image and language tokens.}
  \label{fig:org}
\end{figure*}

\paragraph{Memory-dependent tasks.}
Given an instruction $\ell$, a VLA policy observes $o_t$ (camera images and robot state) at decision $t$ and predicts an action chunk $a_t$.
The history $h_t$ contains all observations and executed actions up to $t$, plus any demonstration given with the task.
A task is \emph{memory-dependent} if two histories with the same instruction and current observation require different expert actions:
\begin{equation}
\pi^\star(\cdot\mid\ell,h_t) \neq \pi^\star(\cdot\mid\ell,h'_t)
\quad \text{with } o_t = o'_t .
\label{eq:memory-dependence}
\end{equation}
Pressing a button three times illustrates this: the scene may look identical after the first and third press, yet the correct action differs. A policy conditioned only on $(\ell, o_t)$ cannot distinguish them.

\begin{figure*}[t]
  \centering
  \includegraphics[width=1\linewidth]{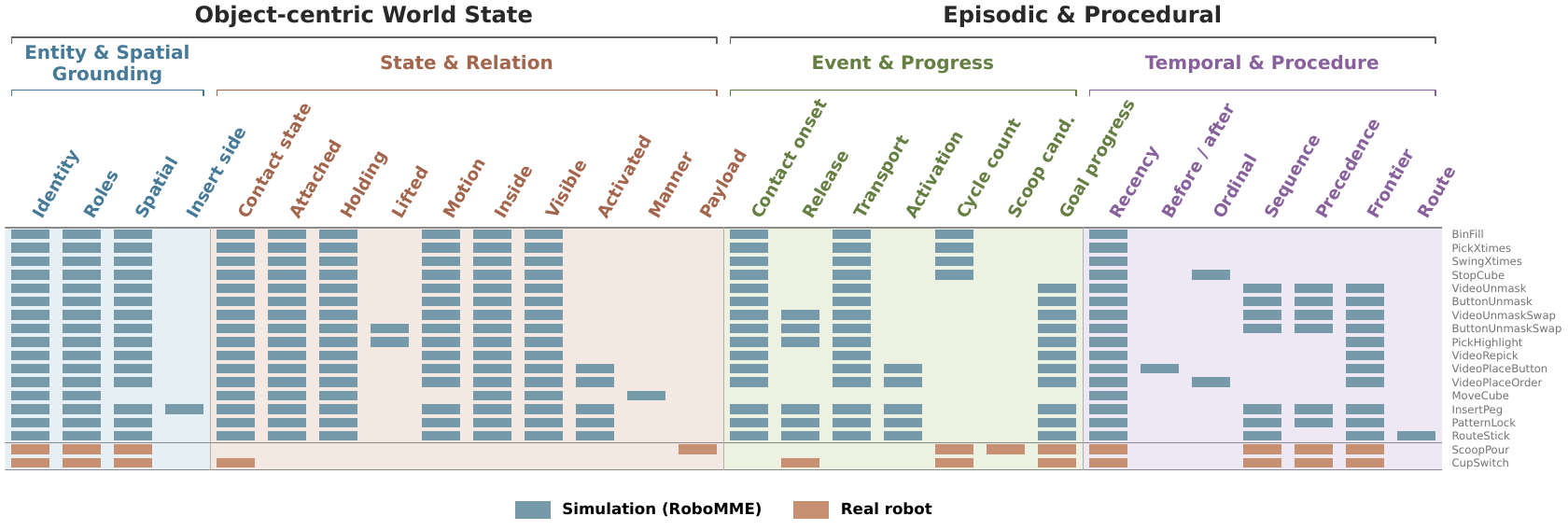}
  \caption{\textbf{A shared concept library supports diverse memory tasks.}
Columns show concepts grouped into four semantic families, and rows show the 16 RoboMME tasks and two real-robot tasks.
Filled cells indicate the concepts used by each task, revealing extensive reuse of shared concepts across otherwise different memory demands.}
  \label{fig:concept-library}
\end{figure*}

\paragraph{Memory-conditioned policy.}
The history is long, but what separates $h_t$ from $h'_t$ is usually a few facts.
We therefore compress the history into a memory $m_t$: a set of concept records, each stating one fact about the task entities (Figure~\ref{fig:concept-library}).
We separate what to remember from how to use it:
\begin{equation}
m_t = W(\ell,h_t), \qquad u_t=R_\phi(m_t), \qquad
a_t \sim \pi_\theta(\cdot\mid\ell,o_t,u_t).
\label{eq:factorization}
\end{equation}
We construct a Writer $W$ that selects concepts, harnesses pretrained perception models to recognize them, and updates records using only the history available at $t$ (Section~\ref{sec:constructing-memory}).
We keep $W$ fixed, because action supervision alone does not specify which facts to keep; a learned memory can instead exploit shortcuts~\citep{causalconfusion,copycat}.
We then train a Reader $R_\phi$ jointly with the VLA $\pi_\theta$ on expert action chunks $a_t^\star$.
The Reader encodes the records into memory tokens $u_t$ that condition the policy (Section~\ref{sec:memory-vla}, Appendix~\ref{app:training}).

\section{\method}
\label{sec:method}

\method adds an explicit memory channel to a VLA (Figure~\ref{fig:org}),
following the factorization in \Eqref{eq:factorization}.
The \emph{Writer} $W$ determines what to remember: it draws from a shared
concept library (Section~\ref{sec:concept-library}), selects the concepts
required by the task, recognizes them from visual and robot-state
evidence, and updates the memory records
(Section~\ref{sec:constructing-memory}).
The \emph{Reader} $R_\phi$ learns how to use this memory by encoding the
records into tokens that condition the VLA
(Section~\ref{sec:memory-vla}).

\subsection{Concept Library}
\label{sec:concept-library}

\paragraph{A shared concept library.}
Memory-dependent tasks differ, but the facts they need recur.
Counting button presses and counting placements both need the number of completed events.
Finding a hidden cube and re-picking a demonstrated object both need a remembered location.
We therefore define each type of fact once, as a concept in a shared library.
A new task reuses these concepts instead of building from scratch, and a single Reader serves all tasks.
Robot knowledge bases and neuro-symbolic concept learners follow the same principle~\citep{tenorth2009knowrob,mao2019nscl}.

\paragraph{Four concept families.}
We keep a concept only if it can change the correct action when the instruction and current observation are the same (\Eqref{eq:memory-dependence}).
The concepts that meet this criterion fall into four families (Figure~\ref{fig:concept-library}).
Each family resolves an ambiguity that the previous ones cannot.
\textbf{Entity \& Spatial Grounding} answers \emph{what and where}.
A remembered fact is useful only if it stays attached to the right object, so this family binds facts to persistent entities and their locations~\citep{gu2024conceptgraphs}.
\textbf{State \& Relation} answers \emph{in what state}.
Knowing the target does not tell the robot what to do with it, so this family records states and relations such as \texttt{holding} and \texttt{inside}~\citep{predicateinvention}.
\textbf{Event \& Progress} answers \emph{what has happened}.
The same scene can call for continuing or stopping, so this family records completed events and progress toward the goal~\citep{huang2023inner}.
\textbf{Temporal \& Procedure} answers \emph{in what order}.
Counts ignore order, so this family records the order of events and the required order of steps.
The first two families describe the current world; the last two describe the past.

\paragraph{Concepts and records.}
We construct each concept with a coding agent. The library is built once and then reused in most of the other tasks.
It specifies its arguments, a recognizer that harnesses pretrained perception models and robot state, and an update rule written as explicit logic.
The library stores concepts; memory stores their instances as records.
For example, the record \texttt{INSIDE(s2, s1)} states that entity \texttt{s2} is inside container \texttt{s1}, with a truth value, a status, a time, and its supporting evidence.
Concepts also compose.
The pick--place concept builds on the transport concept, which in turn builds on contact and region visits.

\paragraph{Extending the library.}
We build the library to grow with new tasks.
Because concepts compose, a new concept reuses the existing perception models, lower-level concepts, and record format.
An agent constructs the new concept on top of the library, and we verify it on a few demonstrations (Appendix~\ref{app:enrichment}).
For example, scooping beans into a bowl twice needs a scooping concept, which the library lacks.
We could add \texttt{SCOOP} to the library: it reuses entity grounding, containment, and event counting, and adds only a check that the scooper carries beans (Section~\ref{sec:real}).

\subsection{Constructing Concept Memory}
\label{sec:constructing-memory}

The Writer $W$ builds the memory $m_t$ in three steps: it selects the concepts a task needs, recognizes them from the history $h_t$, and updates the records.

\paragraph{Task-conditioned selection and grounding.}
Running every concept is costly, and it adds records the task does not need, which distract the policy (Section~\ref{sec:recognition-ablation}).
The Writer therefore activates only the concepts the instruction requires.
A fixed instruction parser returns the required concepts $S(\ell,\mathcal L)$, and $\operatorname{DepClosure}$ adds the concepts they depend on.
For each active concept, $\operatorname{Inst}$ extracts its parameters $\eta_c$ from the instruction:
\begin{equation}
\mathcal C_\ell = \operatorname{DepClosure}\bigl(S(\ell,\mathcal L)\bigr),
\qquad
\eta_c = \operatorname{Inst}(\ell,c), \quad c\in\mathcal C_\ell.
\label{eq:concept-selection}
\end{equation}
The parameters $\eta_c$ include entity phrases for \emph{objects}, \emph{destinations}, \emph{controls}, and also the required properties, such as \emph{counts}, \emph{ordinals}, and \emph{order constraints} (Appendix~\ref{app:library}). These two operations output answers to different questions: $\mathcal C_\ell$ decides which concepts to recognize, and $\eta_c$ decides which entities each concept grounds, so each record binds to the object the task refers to rather than to any object in view.
Task conditioning only decides what to observe but does not establish facts: \emph{repeat three times} sets a desired count, and only observed completion events increase the completed count.

\paragraph{Multimodal concept recognition.}
No single modality recognizes a concept reliably.
Vision finds objects but cannot tell whether the gripper holds one; robot state tells when the gripper closes but not on what.
Each recognizer therefore composes evidence across modalities. We harness OWLv2~\citep{minderer2023owlv2} to ground the entity phrases as bounding boxes, and SAM2~\citep{sam2} to track them across frames where needed.
End-effector positions and gripper states show when and where the robot acts, and camera calibration projects the gripper into the image, so visual and robot-state evidence can be compared directly.
Each active concept $c$ runs its recognizer $\operatorname{Rec}_c(h_t;\eta_c)$, which composes these signals at three levels.
Spatial and relational checks describe how entities are arranged at one time.
Event checks test changes over an interaction interval.
Temporal operators assemble confirmed events into counts or ordered sequences. Each result keeps its entity bindings, evidence, and time.

\paragraph{Evidence merging and memory update.}
Recognition outputs are noisy and repetitive.
The same event appears in many frames, and objects disappear behind occluders.
The Writer merges the outputs of all active concepts and updates the previous memory:
\begin{equation}
m_t = \operatorname{Update}\Bigl(m_{t-1},\ \bigl\{\operatorname{Rec}_c(h_t;\eta_c)\bigr\}_{c\in\mathcal C_\ell}\Bigr).
\label{eq:memory-update}
\end{equation}
$\operatorname{Update}$ turns noisy outputs into stable memory with four rules.
First, a state needs accumulated evidence before it is confirmed, and a missed detection is not evidence against it; an occluded object therefore keeps its last location.
Second, an event is recorded once, when it is first confirmed; seeing it again adds nothing.
Third, a count is the number of distinct recorded events.
Fourth, progress compares counts with the goal, and an ordered step counts only after its prerequisites are complete.
All updates use only observations up to the current decision.
A demonstration given with the task can set entity bindings and the required order, but its events never count as the robot's own progress.
\subsection{Integrating Concept Memory with VLA}
\label{sec:memory-vla}

The Reader $R_\phi$ turns the memory $m_t$ into tokens $u_t$ that the VLA $\pi_\theta$ can attend to.

\paragraph{Reading memory records as tokens.}
Each record $r\in m_t$ is a tuple of $K$ discrete fields: kind, predicate, subject and object entities, truth value, status, confidence, age, progress flags, and grid location (Appendix~\ref{app:memory-details}).
Writing records as text would flatten this structure and lengthen the prompt, so the Reader embeds the fields directly.
Each field $k$ has its own embedding table $E_k$, because the same value means different things in different fields.
Summing the field embeddings gives one vector $e(r)$ per record.
A small Transformer $T_\phi$ then relates the records and maps each one to the VLA's token width:
\begin{equation}
e(r)=\sum_{k=1}^{K}E_k(r_k),\qquad
u_t = R_\phi(m_t) = T_\phi\bigl(\{e(r)\}_{r\in m_t}\bigr).
\label{eq:memory-encoding}
\end{equation}
The Reader's parameters $\phi$ are the tables $E_k$ and the Transformer $T_\phi$.
The Reader outputs one token per record, without pooling, so no fact is averaged away.
Attention across records lets it relate facts, such as a count and its goal.
Records form a set, so the Reader uses no position embeddings; temporal information enters through the age field and goal dependencies.

\paragraph{Policy integration and training.}
The tokens $u_t$ join the image and language tokens in the prefix of $\pi_\theta$, which we initialize from pretrained $\pi_{0.5}$~\citep{pi2025pi05}.
The pretrained policy has never seen memory tokens, so it must learn how they relate to actions.
We therefore train $\phi$ jointly with the language backbone and action expert of $\pi_\theta$, using the base model's flow-matching objective.
The vision encoder and the Writer $W$ stay frozen.
We compute training memory as at execution: $m_t$ uses only observations up to decision $t$.
Entity identifiers are arbitrary labels, so we randomly permute them in each training sample; the policy cannot rely on specific numbers.
At execution, $W$ updates $m_t$ as observations arrive, and $R_\phi$ reads it before each action chunk.

\section{Experiments}
\label{sec:experiments}

Our experiments address four questions.
\textbf{(1)} How does explicit concept memory compare with no-memory, history-based, and visual-memory alternatives (Section~\ref{sec:overall-evaluation})?
\textbf{(2)} Do different concept families provide distinct information needed for memory-dependent decisions (Section~\ref{sec:content-ablation})?
\textbf{(3)} What is required to construct useful concept memory, and how important are structured recognition, task-conditioned selection, and language grounding (Section~\ref{sec:recognition-ablation})?
\textbf{(4)} Can the concept library transfer to new real-robot tasks through reuse and extension (Section~\ref{sec:real})?

\subsection{Experiment Setup}
\label{sec:experiment-setup}

\paragraph{Tasks.}
We evaluate on the 16 memory-dependent tasks of RoboMME~\citep{dai2026robomme} and two new real-robot tasks.
RoboMME covers four memory demands: counting repeated events, maintaining objects through occlusion, referring to past interactions, and reproducing demonstrated behavior.
On the real robot, \textbf{CupSwitch} requires completing a requested number of cup round trips between two plates, while \textbf{ScoopPour} requires a requested number of successful scoop-and-pour cycles; empty scoops do not count.
These tasks introduce new objects and success conditions, and ScoopPour additionally requires a concept absent from the base library.
We use a 7-DoF YAM arm and collect demonstrations with GELLO~\citep{gello}; further platform details are in Appendix~\ref{app:real-robot}.

\paragraph{Baselines.}
On RoboMME, we compare against the baselines reported by RoboMME: $\pi_{0.5}$ without memory, $\pi_{0.5}$ with past actions, FrameSamp+Modul, SAM2Act+, and MemER.
Together they cover no memory, history conditioning, and visual-memory approaches.
We additionally compare the Writer with a fine-tuned VLM that extracts the same concepts directly from video (Section~\ref{sec:recognition-ablation}).
On the real robot, we compare against $\pi_{0.5}$ fine-tuned on the same demonstrations without memory.

\paragraph{Training and evaluation.}
We initialize \method from pretrained $\pi_{0.5}$ and jointly train on all 16 RoboMME tasks for 200,000 updates.
Each task is evaluated on 50 validation episodes with three policy sampling seeds, and we report per-task success and the task-equal mean.
Controlled ablations use matched 50-episode evaluations and their own full-memory controls; their absolute scores are therefore not directly comparable with Table~\ref{tab:main-results} (Appendices~\ref{app:training} and~\ref{app:evaluation}).
For the real robot, both methods are fine-tuned from pretrained $\pi_{0.5}$ on the same teleoperated demonstrations, and success requires completing exactly the requested number of cycles (Appendix~\ref{app:real-robot}).

\begin{table}[!t]
\caption{\textbf{Success rates (\%) on the 16 RoboMME memory-dependent tasks.}
AVG is the task-equal mean across all tasks.
Baseline and human results are reported by RoboMME~\citep{dai2026robomme}.}
\label{tab:main-results}
\centering

\setlength{\tabcolsep}{2.5pt}
\renewcommand{\arraystretch}{1.2}

\resizebox{\textwidth}{!}{%
\begin{tabular}{l|cccc|cccc|cccc|cccc|c}

\multicolumn{1}{c|}{\bf Method}
&
\multicolumn{4}{c|}{\bf Counting}
&
\multicolumn{4}{c|}{\bf Permanence}
&
\multicolumn{4}{c|}{\bf Reference}
&
\multicolumn{4}{c|}{\bf Imitation}
&
\multicolumn{1}{c}{\bf AVG}
\\

\cline{2-5}
\cline{6-9}
\cline{10-13}
\cline{14-17}

&
\shortstack{\bf Bin\\\bf Fill}
&
\shortstack{\bf Pick\\\bf Xtimes}
&
\shortstack{\bf Swing\\\bf Xtimes}
&
\shortstack{\bf Stop\\\bf Cube}
&
\shortstack{\bf Video\\\bf Umsk}
&
\shortstack{\bf Button\\\bf Umsk}
&
\shortstack{\bf Video\\\bf UmskS}
&
\shortstack{\bf Button\\\bf UmskS}
&
\shortstack{\bf Pick\\\bf HighL}
&
\shortstack{\bf Video\\\bf Repick}
&
\shortstack{\bf Video\\\bf PicBtn}
&
\shortstack{\bf Video\\\bf PicOrd}
&
\shortstack{\bf Move\\\bf Cube}
&
\shortstack{\bf Insert\\\bf Peg}
&
\shortstack{\bf Pattern\\\bf Lock}
&
\shortstack{\bf Route\\\bf Stick}
&
\\

\hline

Human Performance
& \textcolor{blue!60!black}{\textbf{96.00}}
& \textcolor{blue!60!black}{\textbf{100.00}}
& \textcolor{blue!60!black}{\textbf{80.00}}
& \textcolor{blue!60!black}{\textbf{78.00}}
& \textcolor{blue!60!black}{\textbf{90.00}}
& \textcolor{blue!60!black}{\textbf{92.00}}
& \textcolor{blue!60!black}{\textbf{92.00}}
& \textcolor{blue!60!black}{\textbf{90.00}}
& \textcolor{blue!60!black}{\textbf{92.00}}
& \textcolor{blue!60!black}{\textbf{92.00}}
& \textcolor{blue!60!black}{\textbf{98.00}}
& \textcolor{blue!60!black}{\textbf{90.00}}
& \textcolor{blue!60!black}{\textbf{90.00}}
& \textcolor{blue!60!black}{\textbf{98.00}}
& \textcolor{blue!60!black}{\textbf{84.00}}
& \textcolor{blue!60!black}{\textbf{86.00}}
& \textcolor{blue!60!black}{\textbf{90.50}}
\\

FrameSamp+Modul
& 39.56
& 87.33
& 92.00
& 42.00
& 32.67
& 25.11
& 24.44
& 18.22
& 22.89
& 30.44
& 60.00
& 32.00
& 77.78
& 7.56
& 53.56
& 66.67
& 44.51
\\

$\pi_{0.5}$
& 30.00
& 42.89
& 35.56
& 6.67
& 20.44
& 22.22
& 18.67
& 6.67
& 11.33
& 0.44
& 31.11
& 25.78
& 26.00
& 1.56
& 2.89
& 4.67
& 17.93
\\

$\pi_{0.5}$ w/ past actions
& 26.67
& 58.33
& 26.67
& 4.67
& 30.67
& 23.67
& 20.67
& 16.00
& 12.33
& 8.67
& 24.00
& 18.67
& 34.00
& 1.00
& 4.00
& 5.67
& 19.73
\\

SAM2Act+
& 40.00
& 76.00
& 25.33
& 0.00
& 27.33
& 32.00
& 18.00
& 26.67
& 17.33
& 5.33
& 24.67
& 20.00
& 29.33
& 0.00
& 0.00
& 0.00
& 21.37
\\

MemER
& 56.67
& 79.33
& 59.33
& 0.00
& 81.33
& 72.00
& 38.00
& 21.33
& \textcolor{red!70!black}{\textbf{70.67}}
& 25.33
& 30.00
& 26.00
& 82.67
& 6.67
& 16.67
& 12.00
& 42.38
\\

\hline

\textbf{ECoMEM}
& \textcolor{red!70!black}{\textbf{71.33}}
& \textcolor{red!70!black}{\textbf{100.00}}
& \textcolor{red!70!black}{\textbf{99.33}}
& \textcolor{red!70!black}{\textbf{66.00}}
& \textcolor{red!70!black}{\textbf{100.00}}
& \textcolor{red!70!black}{\textbf{96.67}}
& \textcolor{red!70!black}{\textbf{78.00}}
& \textcolor{red!70!black}{\textbf{84.00}}
& \textbf{62.67}
& \textcolor{red!70!black}{\textbf{77.33}}
& \textcolor{red!70!black}{\textbf{96.00}}
& \textcolor{red!70!black}{\textbf{86.00}}
& \textcolor{red!70!black}{\textbf{96.00}}
& \textcolor{red!70!black}{\textbf{40.67}}
& \textcolor{red!70!black}{\textbf{79.33}}
& \textcolor{red!70!black}{\textbf{85.33}}
& \textcolor{red!70!black}{\textbf{82.42}}
\\

\hline

\end{tabular}%
}

\end{table}

\subsection{Concept Memory Outperforms Implicit and Visual Memory}
\label{sec:overall-evaluation}

\textbf{Explicit concept memory substantially improves memory-dependent control.}
\method achieves 82.42\% average success, 37.91 points above the strongest baseline, FrameSamp+Modul, and leads the evaluated policies on 15 of 16 tasks (Table~\ref{tab:main-results}).
Its average is within 8.08 points of human performance.

\textbf{The advantage is largest when the required information must persist beyond the visible frame.}
The largest margins occur on ButtonUnmaskSwap, VideoPlaceOrder, and VideoRepick, where the robot must act on a target, location, or ordering established earlier but no longer directly observable.
Here, \method preserves entity bindings, locations, and event order as persistent records rather than requiring the policy to recover them from visual history.
Counting tasks similarly benefit from explicit completion events and progress.
By contrast, gains are smaller on SwingXtimes and MoveCube, where sampled or retrieved frames already expose much of the relevant history.
PickHighlight is the only task where \method does not lead: MemER is 8 points higher, consistent with a transient visual cue being directly preserved in a retrieved keyframe.

\textbf{Remaining errors expose control limits beyond memory.}
On InsertPeg, \method reaches 40.67\%, compared with 7.56\% for the strongest baseline, but remains well below human performance.
Memory provides the insertion side and endpoint; precise execution must still be learned by the policy.
Section~\ref{sec:content-ablation} next isolates which information in concept memory drives these gains.

\begin{figure}[!htbp]
\begin{minipage}[t]{0.32\textwidth}
  \vspace{0pt}
  \centering
\includegraphics[width=\linewidth,height=0.83\linewidth,keepaspectratio,trim=0 34 0 24,clip]{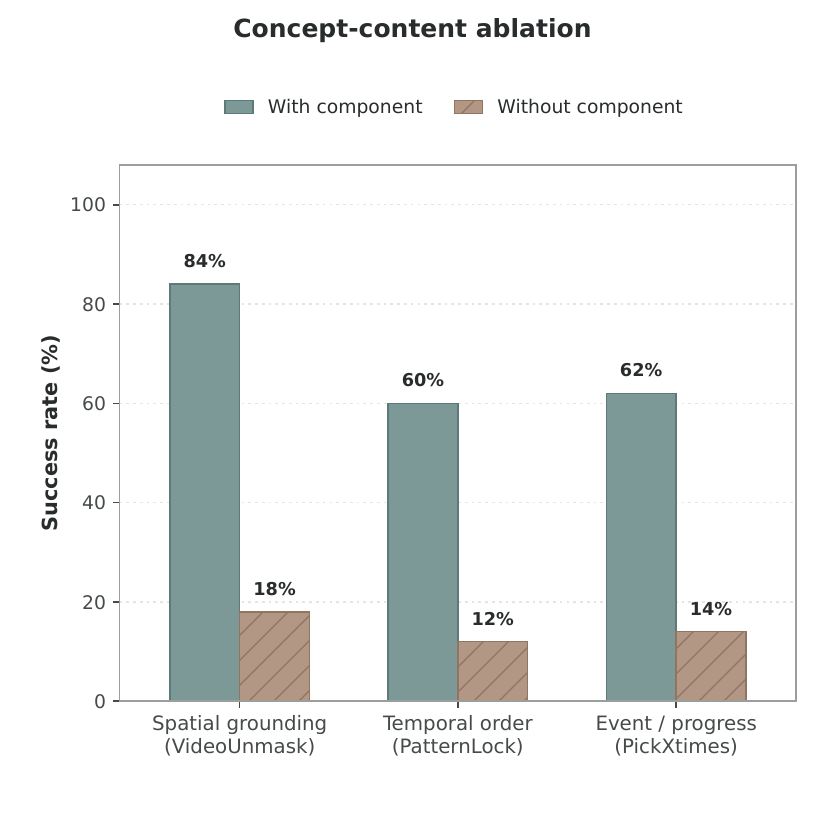}
  \caption{\textbf{Different memory decisions depend on complementary concept families.}
Removing spatial grounding, temporal order, or event and progress information sharply reduces success on the corresponding memory-dependent task.
Each pair is trained and evaluated with the same intervention on 50 matched episodes.}
  \label{fig:content-ablations}
\end{minipage}\hfill
\begin{minipage}[t]{0.32\textwidth}
  \vspace{0pt}
  \centering
  \includegraphics[width=\linewidth,height=0.83\linewidth,keepaspectratio,trim=0 34 0 24,clip]{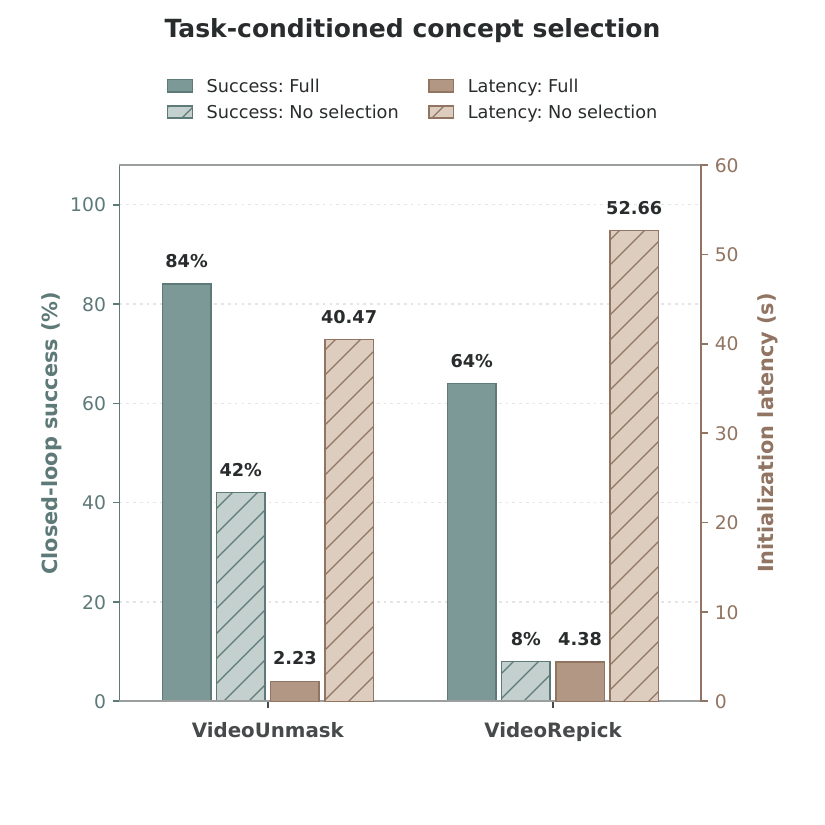}
  \caption{\textbf{Task-conditioned concept selection improves both control and efficiency.}
With the policy fixed, full-library recognition adds task-unselected records, reduces success on VideoUnmask and VideoRepick, and increases memory initialization latency by $12.0\times$ to $18.1\times$.
Latency is measured on matched episodes using the same GPUs.}
  \label{fig:concept-selection}
\end{minipage}\hfill
\begin{minipage}[t]{0.32\textwidth}
  \vspace{0pt}
  \centering
  \includegraphics[width=\linewidth,height=0.83\linewidth,keepaspectratio,trim=0 34 0 24,clip]{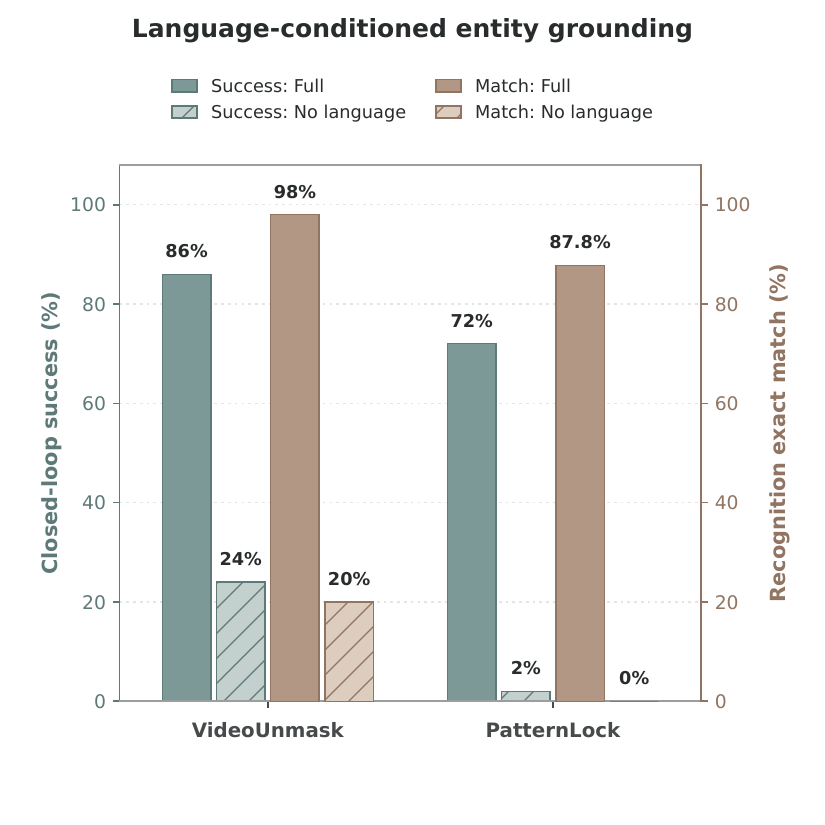}
  \caption{\textbf{Language-conditioned grounding binds memory to the correct entities.}
Replacing instruction-specific entity phrases with generic role-level queries sharply reduces both reference-record agreement and closed-loop success on VideoUnmask and PatternLock.
Concept selection and the policy are fixed.}
  \label{fig:language-grounding}
\end{minipage}
\end{figure}

\subsection{Concept Families Provide Complementary Memory}
\label{sec:content-ablation}

\textbf{A compact library is shared across diverse memory tasks.} As shown in Figure~\ref{fig:concept-library}, the 16 RoboMME tasks reuse concepts from the same four families rather than requiring task-specific memory representations. Tasks may differ in which concepts they activate and how they compose them.

\textbf{Different memory decisions depend on different concept families.}
We remove spatial information on VideoUnmask, event and progress information on PickXtimes, and temporal order on PatternLock, while preserving the remaining memory and removing indirect cues to the ablated information (Appendix~\ref{app:content-eval}).
Each intervention is applied during both training and evaluation on matched data, allowing the policy to exploit any information that remains.
Success drops from 84\% to 18\%, 62\% to 14\%, and 60\% to 12\%, respectively (Figure~\ref{fig:content-ablations}; exact McNemar $p<0.002$ for all three).
The failures align with the role of each family: \emph{where} to act after occlusion, \emph{how many} events have occurred, and \emph{in what order} to act.
Together, the coverage in Figure~\ref{fig:concept-library} and these targeted ablations show that a small shared vocabulary can support diverse memory demands while preserving the distinctions needed for control.

\begin{table}[t]
\centering
\caption{\textbf{Concept extraction across all four memory families.}
We evaluate eight whole-video concepts on 95 RoboMME test videos.
Structured and QA are separately fine-tuned Qwen3-VL-8B-Instruct models;
Concept Memory uses our evidence-based Writer.
Cell shading encodes extraction accuracy (darker is higher);
\textbf{bold} denotes the best result in each row.
See Appendix~\ref{app:vlm_concept_extraction} for protocols and limitations.}
\label{tab:vlm_concept_extraction}

\small
\setlength{\tabcolsep}{6pt}
\renewcommand{\arraystretch}{1.15}

\begin{tabular}{llccc}
\toprule
Family & Concept extracted
& Structured & QA & \textbf{Concept Memory} \\
\midrule

\multirow{2}{*}{Entity \& Spatial}
& Picked-object identity
& \score{71.4}
& \score{71.4}
& \scoreb{100.0} \\

& Highlighted-object set
& \score{63.2}
& \score{63.2}
& \scoreb{97.4} \\

\cmidrule(lr){1-5}

\multirow{2}{*}{State \& Relation}
& Currently highlighted
& \scoreb{100.0}
& \score{97.4}
& \score{97.4} \\

& Currently holding queried object
& \scoreb{100.0}
& \score{96.7}
& \scoreb{100.0} \\

\cmidrule(lr){1-5}

\multirow{2}{*}{Event \& Progress}
& Pick--place cycle count
& \score{86.7}
& \score{93.3}
& \scoreb{100.0} \\

& Region-entry count
& \scoreb{95.0}
& \score{90.0}
& \scoreb{95.0} \\

\cmidrule(lr){1-5}

\multirow{2}{*}{Temporal \& Procedure}
& Ordered object--destination sequence
& \score{11.8}
& \score{5.9}
& \scoreb{100.0} \\

& Placements before/after press
& \score{23.5}
& \score{23.5}
& \scoreb{88.2} \\

\bottomrule
\end{tabular}
\end{table}

\subsection{Reliable Memory Requires Structured Construction}
\label{sec:recognition-ablation}

The Writer must do more than preserve history: it must extract the right
structure, select what is relevant, and bind records to the right
entities. We test each requirement against a simpler alternative.
For structured extraction, we compare Concept Memory with fine-tuned
VLMs on concepts spanning all four memory families. For selection and
grounding, we intervene on the Writer at evaluation while keeping the
trained policy fixed and using matched full-memory controls.

\textbf{Direct VLM extraction is strong on local state, but weak on
structured history.}
We fine-tune Qwen3-VL-8B-Instruct to extract each concept directly from
video, either as a structured record or as an answer to a question, and
evaluate eight concepts on 95 held-out RoboMME videos
(Table~\ref{tab:vlm_concept_extraction}).
The VLMs recover current state and simple event counts reliably:
state-and-relation concepts reach 96.7--100\%, and event counts reach
86.7--95.0\%.
Performance drops for episodic entity information
(63.2--71.4\%) and collapses on temporal composition, reaching only
5.9--11.8\% for ordered object--destination sequences and 23.5\% for
placements around an event.
The similar Structured and QA results indicate that output format is not
the main bottleneck.
In contrast, Concept Memory achieves 95--100\% on the first three
families and 88.2--100\% on temporal concepts.
Explicitly composing grounded evidence into events and temporal records
therefore provides a substantially more reliable memory construction
mechanism than the tested direct video extraction.

\textbf{Task-conditioned selection improves both accuracy and efficiency.}
\emph{No Selection} runs every recognizer and adds all resulting records
while retaining the task-relevant ones.
Success drops from 84\% to 42\% on VideoUnmask and from 64\% to 8\% on
VideoRepick, while median memory initialization becomes $18.1\times$ and
$12.0\times$ slower (Figure~\ref{fig:concept-selection}).
Thus, more memory is not necessarily better: task-conditioned selection
reduces recognition cost and prevents irrelevant records from disrupting
the policy.

\textbf{Without language grounding, records bind to the wrong objects.}
\emph{Generic queries} replace task-specific entity phrases with
role-level ones such as \emph{an object}, while keeping the selected
concepts, detection budget, and policy instruction fixed.
The Writer runs the right concepts but no longer knows which entities
they refer to.
Agreement with reference records falls from 98\% to 20\% for target
locations in VideoUnmask and from 87.8\% to 0\% for ordered location
sequences in PatternLock; closed-loop success falls from 86\% to 24\%
and from 72\% to 2\% (Figure~\ref{fig:language-grounding}).
The policy still receives the full instruction, yet it cannot repair
memory records grounded to the wrong objects.

Together, these results show that useful memory depends on how it is
constructed. Reliable memory requires structured evidence composition,
task-conditioned selection, and language-grounded entity binding; the
policy cannot reliably recover information that the Writer constructs
incorrectly.

\begin{table}[t]
\centering
\small
\setlength{\tabcolsep}{6pt}
\caption{Real-robot success (successes/trials) by requested number of cycles; success requires exactly that count. Pooled uses the targets both policies were run on.}
\label{tab:real_robot}
\begin{tabular}{lccccccc}
\toprule
 & \multicolumn{3}{c}{ScoopPour} & \multicolumn{2}{c}{CupSwitch} & \multicolumn{2}{c}{Pooled} \\
\cmidrule(lr){2-4}\cmidrule(lr){5-6}\cmidrule(lr){7-8}
Method & $1\times$ & $2\times$ & $3\times$ & $2\times$ & $3\times$ & succ./trials & rate \\
\midrule
ECoMEM                   & 16/20 & 19/23 & 11/12 & 11/12 & 11/12 & \textbf{68/79} & \textbf{86.1\%} \\
$\pi_{0.5}$ (no memory)  & 4/12  & 0/12  & 0/11  & 0/12  & 1/11  & 5/58  & 8.6\% \\
\bottomrule
\end{tabular}
\end{table}

\begin{figure*}[t]
\centering
\includegraphics[width=\linewidth]{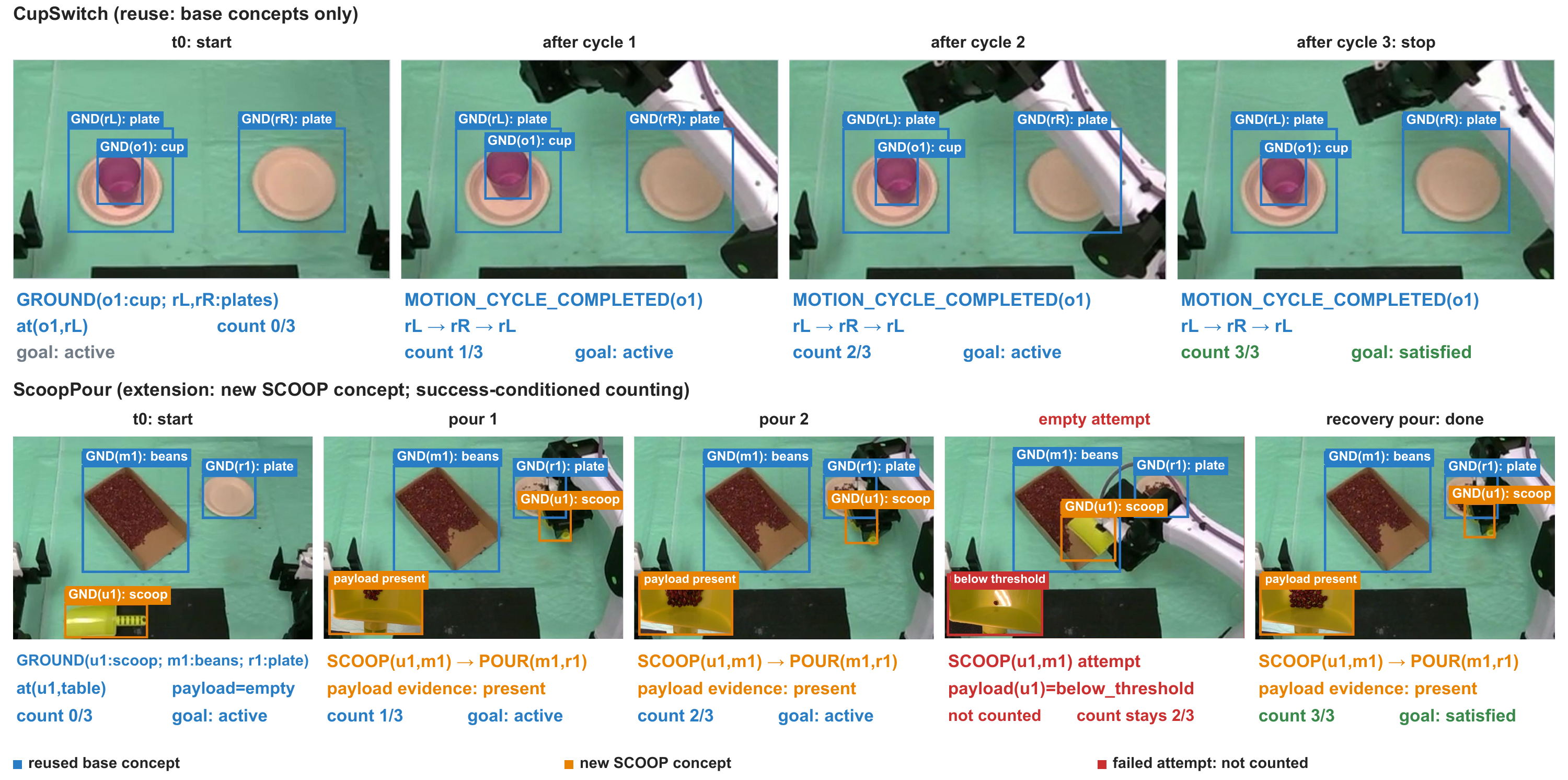}
\caption{\textbf{Concept memory transfers through reuse and compositional extension on a real robot.}
\textcolor{reusedc}{Blue}: reused base concepts; \textcolor{newc}{orange}: the new \texttt{SCOOP} concept.
Both tasks require counting, but \method represents progress at the task-relevant granularity: CupSwitch advances after each completed round trip, while ScoopPour composes \texttt{SCOOP} with the same counter so that only successful scoops carrying beans advance progress.
An empty scoop therefore leaves the count unchanged and automatically triggers a retry.
Records are produced online by the Writer.}
\label{fig:real_qual}
\end{figure*}

\subsection{The Library Transfers to New Tasks on a Real Robot}
\label{sec:real}
\label{sec:library-enrichment}

The real robot tests whether the library transfers beyond RoboMME to new cameras, objects, and tasks.
We consider two forms of transfer: \textbf{CupSwitch} tests whether a new task can reuse the existing library, while \textbf{ScoopPour} tests whether a missing capability can be added compositionally.
Both policies initialize from pretrained $\pi_{0.5}$ and use the same real-robot demonstrations; \method additionally receives concept memory.

\textbf{Transfer occurs through reuse and local extension.}
CupSwitch uses only existing concepts for grounding, persistence, pick-place cycles, event counting, and count goals, with no new recognizer or update rule (Figure~\ref{fig:concept-library}).
ScoopPour adds a single concept, \texttt{SCOOP}(tool, material), which confirms a scoop only when the lifted scooper carries beans.
Grounding, persistence, counting, and goal tracking are reused unchanged.
Thus, one new task is covered directly, while the other extends the library with one concept that composes with the existing machinery.

\textbf{Conditioning on explicit memory is critical for repeated tasks.}
\method succeeds in 80\% to 92\% of ScoopPour trials and 92\% of CupSwitch trials, remaining high across requested counts (Table~\ref{tab:real_robot}).
Without memory, $\pi_{0.5}$ reaches 33\% on one ScoopPour cycle but 0\% on two and three, and reaches 0\% and 9\% on two and three CupSwitch round trips.
Across shared evaluation conditions, \method succeeds in 68/79 trials (86.1\%), compared with 5/58 (8.6\%) for the no-memory policy (Fisher exact $p<10^{-20}$; trials are unpaired and unequal in number).
The two systems share the pretrained backbone and demonstrations, while only \method conditions the policy on explicit memory tokens.
The collapse of the no-memory policy as counts increase shows that current observations alone are insufficient once task progress is no longer visible.

\textbf{Neural-symbolic concepts let memory adapt its granularity to the task.}
Because concepts compose learned perceptual recognition with explicit event and progress structure, the same memory machinery can track different notions of task progress.
Figure~\ref{fig:real_qual} shows this online.
In CupSwitch, the counter advances after each completed round trip, distinguishing visually similar scenes by how many cycles have occurred.
In ScoopPour, \texttt{SCOOP} adds a success condition, so only scoops that actually carry beans advance the same counter; an empty scoop leaves progress unchanged and the policy retries.
Thus, the library can represent progress at the semantic granularity each task requires by composing shared counting machinery with task-relevant concepts.

\section{Conclusion}
\label{sec:conclusion}
We introduced \method, an explicit concept memory interface that separates evidence-based memory construction from learned action generation.
A shared, compositional concept library represents task-relevant world state, events, progress, and temporal structure, while the Writer maintains these records and the Reader learns how to use them for control.
Across 16 RoboMME tasks, \method substantially outperforms no-memory, history-based, and visual-memory alternatives.
Controlled ablations show that distinct concept families provide complementary information, and that structured recognition, task-conditioned selection, and language grounding are critical to constructing effective memory.
On new real-robot tasks, the same library transfers through direct reuse or the addition of a single task-specific concept, showing that explicit concepts can provide a reusable and extensible memory interface for VLA control.

\clearpage
\bibliography{references}
\bibliographystyle{plainnat}

\clearpage
\appendix
\section{Policy Training}
\label{app:training}

\subsection{Shared policy configuration.}
All policies use pretrained $\pi_{0.5}$~\cite{pi2025pi05}, a
2B language backbone, and a 300M action expert. The image backbone is
frozen; the language backbone, action expert, and memory reader are
optimized jointly. Recognition and memory writing remain fixed.
All policy training is performed on NVIDIA H200 GPUs.
The ECoMEM policy predict 20 actions and execute
the first 16 before requesting another chunk.

\subsection{Main benchmark training.}
We initialize ECoMEM from pretrained $\pi_{0.5}$ and jointly train on
all 16 RoboMME tasks. Samples are drawn uniformly from the
retained observation--action frames.
Reported results use a single EMA checkpoint. The training process takes $4 \times$ H200 GPU for 24 hours.

\subsection{Controlled comparisons.}
Each content-ablation pair is trained independently from the same
pretrained $\pi_{0.5}$ policy.
The intervention is applied identically in training and evaluation.
Each pair has identical observation/action indices
and normalization.
Each experiment retains its own matched full-memory control.
Training budgets and memory-construction settings differ across the
main benchmark and the controlled comparisons.

\subsection{Flow-matching objective.}
Let $\mathcal D$ contain examples $(\ell,o_t,m_t,a_t^\star)$, with
instruction $\ell$, current observation $o_t$, causal memory $m_t$,
and expert action chunk $a_t^\star$ at decision $t$
(Section~\ref{sec:prelim}).
For each sample, $\sigma$ is a uniformly sampled bijection of its
active entity identifiers, applied consistently to both subject and
object fields. Identifier 0 is fixed. This renames entities, not record
positions. We sample $b\sim\mathrm{Beta}(1.5,1)$ and set
$\tau=0.999b+0.001$; independently, $\epsilon$ is standard Gaussian
noise with the action chunk's shape.
Writing $R_\phi$ for the Reader and $v_\theta$ for the policy's
conditional velocity predictor gives
\begin{equation}
\begin{aligned}
a_t^{(\tau)}&=(1-\tau)a_t^\star+\tau\epsilon,\\
\hat v_t&=v_\theta\!\left(a_t^{(\tau)},\tau\mid
\ell,o_t,R_\phi(\sigma(m_t))\right).
\end{aligned}
\label{eq:flow-prediction}
\end{equation}
The optimized parameters are the Reader weights $\phi$ and the
unfrozen policy weights $\theta$:
\begin{equation}
\min_{\phi,\theta}\;
\mathbb E_{\mathcal D,\sigma,b,\epsilon}
\left[\left\|\hat v_t-(\epsilon-a_t^\star)\right\|_2^2\right].
\label{eq:objective}
\end{equation}
The squared error is averaged over action-chunk entries.

\section{Main Benchmark Evaluation}
\label{app:evaluation}

Baseline scores are taken from Table 3 of the first public
RoboMME manuscript~\cite{dai2026robomme}. The descriptions below
summarize the corresponding methods and benchmark adaptations.

\subsection{Current-observation and past-action baselines.}
The standard $\pi_{0.5}$ conditions action generation on the current
images, robot state, and task instruction, without historical visual
inputs. The past-action variant additionally concatenates previous
actions with the language tokens. Its historical information therefore
describes issued actions rather than the visual outcomes of those
actions.

\subsection{Symbolic memory: SimpleSG and GroundSG.}
An auxiliary VLM predicts the next subgoal from the current image and
accumulated subgoal history. SimpleSG expresses this subgoal as a
language command; GroundSG additionally includes image coordinates
to identify the intended target. The predicted subgoal is appended
to the task instruction for $\pi_{0.5}$ to execute.
RoboMME evaluates prompted Gemini-2.5-Pro, Qwen3-VL-4B fine-tuned
on subgoal annotations, and simulator-provided oracle subgoals.

\subsection{Perceptual memory: FrameSamp and TokenDrop.}
Both methods retain historical features extracted by the vision
encoder. FrameSamp uniformly samples frames and spatially pools
their visual tokens. TokenDrop instead ranks image patches by
temporal RGB differences and retains informative patches under
a fixed token budget. It preserves the first frame and uses
spatiotemporal positional encoding. Thus, FrameSamp selects
whole observations, whereas TokenDrop selects changing regions
across observations.

\subsection{Recurrent memory: RMT and TTT.}
These methods summarize incoming visual features through recurrent
updates rather than retaining a growing collection of past frames.
RMT maintains a fixed set of memory slots. At each update, attention
jointly processes the previous slots and a chunk of incoming tokens;
the updated slot representations are carried to the next chunk and
subsequently supplied to the policy.

TTT instead stores historical information in the fast weights of a
small learned module. Incoming features produce training targets
for a local self-supervised update, after which the updated weights
transform query features into history-conditioned outputs.
The fast weights are carried across chunks, while the resulting
features provide the memory representation used for action
generation. RMT therefore updates a token-valued state, whereas
TTT updates a parameter-valued state.

\subsection{Memory integration.}
Each perceptual or recurrent representation is evaluated with three
integration mechanisms. \emph{Context} places memory tokens alongside
the observation and instruction tokens for processing by the VLM.
\emph{Modul} lets action features query memory through cross-attention;
the retrieved features generate scale and shift parameters that
modulate normalized action features before the feed-forward layers.
\emph{Expert} processes memory through a separately parameterized
transformer branch. The action expert attends to both this branch
and the VLM branch, while memory and VLM features remain separate.
These variants distinguish joint input processing, feature
modulation, and a dedicated memory-processing pathway.

\subsection{MemER.}
MemER~\cite{sridhar2026memer} uses a hierarchical policy with persistent
keyframe memory. A high-level VLM selects task-relevant historical
frames and combines them with recent observations to generate language
subtasks. A low-level VLA executes these subtasks using current sensory
inputs. Keyframe filtering consolidates repeated selections, retaining
salient observations across long interactions. RoboMME adapts MemER
using fine-tuned Qwen3-VL-4B and the GroundSG policy.

\subsection{SAM2Act+.}
SAM2Act+~\cite{fang2025sam2act} augments a multi-view manipulation policy
with a memory encoder, memory bank, and memory attention. The encoder
stores visual features together with spatial action information;
memory attention retrieves this context when predicting subsequent
actions. This supports recalling previously observed locations and
interactions. In RoboMME, predicted keyframe actions are executed
by a motion planner, and demonstration videos prefill the memory bank.

\section{Concept Library}

\label{app:library}
\paragraph{Complete library organized by semantic family.}
Table~\ref{tab:concept-inventory} specifies the remembered concepts and
their representations. It covers all named base predicate, event,
and role codes, together with entity bindings, spatial fields, counts,
and temporal structures. Padding and unknown-value codes are encoding
sentinels, not concepts.
Arguments below name semantic participants rather than serialization
columns: entity, gripper, container, region, control, or goal.
Each instance uses episode-local entity
references; confidence, evidence, and time accompany the applicable
records as specified in Appendix~\ref{app:memory-details}.
\begingroup
\small
\newcolumntype{V}[1]{>{\raggedright\arraybackslash}p{#1}}
\setlength{\tabcolsep}{4pt}
\setlength{\LTcapwidth}{\linewidth}
\begin{longtable}{@{}V{0.43\linewidth}V{\dimexpr0.57\linewidth-2\tabcolsep\relax}@{}}
\caption{Semantic inventory of the base Concept Memory interface, grouped
into four families. Monospaced names are record codes; italicized entries
are fields or derived structures. Predicate arguments name the subject
and, where applicable, the object. Gripper identity is stored separately
as agent metadata.}
\label{tab:concept-inventory}\\
\hline
Concept / arguments & Meaning and representation\\
\hline
\endfirsthead
\multicolumn{2}{l}{\small Table~\thetable\ continued}\\
\hline
Concept / arguments & Meaning and representation\\
\hline
\endhead
\hline
\endfoot
\multicolumn{2}{l}{\textbf{Entity \& Spatial Grounding}}\\*
\hline
\emph{Entity identity and language binding}
& Episode-local entity keys link observations, relations, and events.
Groundings retain instruction phrases and aliases; the policy interface
encodes entity slots rather than alias strings.\\
\texttt{role:cube}, \texttt{role:container},
\texttt{role:object}, \texttt{role:region}, \texttt{role:robot}
& The five grounding-role codes in the base policy vocabulary. Each
grounding records an entity key, role, aliases, confidence, and source.\\
\emph{Entity location and endpoint anchor}
& Grounded entity or tool-endpoint locations are supplied to the Reader
as row and column cells on a $32\times32$ image grid. Cell indices are
1-based; zero denotes an unspecified position. Selected goal records
also carry the subject's target cell.\\
\texttt{occupies\_region}(entity, region)
& Current occupancy of a grounded region. Entry and exit are represented
by separate transition events.\\
\texttt{side:left}, \texttt{side:right}
& Entity-bound categorical insertion-side predicates, distinct from
the demonstrated endpoint's spatial anchor.\\
\hline
\multicolumn{2}{l}{\textbf{State \& Relation}}\\*
\hline
\texttt{gripper\_contact}(entity)
& Supported contact with the gripper. The interaction writer stores
the gripper in agent metadata, not as the predicate's object.\\
\texttt{attached}(entity)
& Supported attachment to the gripper, with independent physical
evidence beyond gripper closure.\\
\texttt{holding}(entity)
& Supported control of an entity by the gripper; lifting is represented
separately.\\
\texttt{lifted}(entity)
& The entity has risen from its support.\\
\texttt{object\_motion}(entity)
& Supported object motion, including transport during interaction.\\
\texttt{inside}(entity, container)
& Containment relative to a grounded container.\\
\texttt{visible}(entity)
& Supported visibility. Missing evidence alone does not establish
a confirmed negative state.\\
\texttt{activated}(control)
& Activation state of a grounded button or control.\\
\texttt{inserted}(entity, container)
& Insertion relation between a manipulated entity and its destination.\\
\texttt{release}(entity, destination)
& Supported release; the destination is stored when provided.\\
\texttt{manner:peg\_push},
\texttt{manner:gripper\_push},
\texttt{manner:grasp\_putdown}
& Entity-bound categorical predicates for demonstrated tool-mediated
push, gripper push, or grasp-and-put-down.\\
\hline
\multicolumn{2}{l}{\textbf{Event \& Progress}}\\*
\hline
\texttt{CONTACT\_ACQUIRED}, \texttt{CONTACT\_RELEASED}
& Confirmed positive and negative transitions of \texttt{gripper\_contact}.\\
\texttt{ATTACHMENT\_ESTABLISHED}, \texttt{ATTACHMENT\_CLEARED}
& Confirmed positive and negative transitions of \texttt{attached}.\\
\texttt{HOLDING\_ESTABLISHED}, \texttt{HOLDING\_CLEARED}
& Confirmed positive and negative transitions of \texttt{holding}.\\
\texttt{TRANSPORT\_STARTED}, \texttt{TRANSPORT\_ENDED}
& Confirmed positive and negative transitions of \texttt{object\_motion}.\\
\texttt{CONTAINMENT\_ESTABLISHED}, \texttt{CONTAINMENT\_CLEARED}
& Confirmed positive and negative transitions of \texttt{inside}.\\
\texttt{CONTROL\_ACTIVATED}
& A newly confirmed positive \texttt{activated} state.\\
\texttt{INSERTION\_ESTABLISHED}, \texttt{INSERTION\_CLEARED}
& Confirmed positive and negative transitions of \texttt{inserted}.\\
\texttt{REGION\_ENTERED}, \texttt{REGION\_EXITED}
& Confirmed positive and negative transitions of \texttt{occupies\_region}.\\
\texttt{RELEASE\_CONFIRMED}
& A newly confirmed positive \texttt{release} state.\\
\texttt{VISIBILITY\_ACQUIRED}, \texttt{VISIBILITY\_LOST}
& Confirmed positive and negative transitions of \texttt{visible}.\\
\texttt{motion\_cycle\_complete}(unit, label),
\texttt{MOTION\_CYCLE\_COMPLETED}
& Completion predicate and derived event for a distinct recognized unit.
The progress writer creates one unit entity per completion; its object
binding identifies the operation or category.\\
\texttt{procedure\_milestone\_complete}(subject, object),
\texttt{PROCEDURE\_MILESTONE\_COMPLETED}
& Completion predicate and derived event for a procedural milestone;
the object binding is optional.\\
\emph{Observed repetition count}
& Distinct completed-unit records encode accumulated progress. The
field-based Reader receives these records, not a separate scalar count
column; category-specific counts follow from their object bindings.\\
\emph{Goal satisfaction}
& Goal records specify desired grounded predicates. Progress marks
goals as satisfied, executable frontier, or blocked. Required repetition
counts are task parameters, separate from observed completions.\\
\hline
\multicolumn{2}{l}{\textbf{Temporal \& Procedure}}\\*
\hline
\emph{Event time and evidence provenance}
& Event records store occurrence time, evidence IDs, source, and
confidence. The Reader receives a quantized relative age rather than
an absolute timestamp or evidence-ID string.\\
\emph{Ordered sequence}
& Ordered bindings, goal nodes, and serialized records preserve
demonstrated entity, control, or waypoint sequences. Sequence is a
composed structure, not a separate \texttt{ORDERED\_SEQUENCE} token
in the base field vocabulary.\\
\emph{Ordinal and before/after reference}
& Language parameters specify an ordinal, reference event, and temporal
relation. Recognition resolves them into the corresponding observed
entity or destination.\\
\texttt{route:left\_clockwise},
\texttt{route:left\_counterclockwise},
\texttt{route:right\_clockwise},
\texttt{route:right\_counterclockwise}
& Categorical route predicates attached to ordered hop entities, with
associated waypoint groundings. Route belief records retain hop order.\\
\emph{Goal prerequisites and active step}
& Goal nodes retain prerequisite IDs and current-state or ever-achieved
satisfaction modes. The writer derives frontier and blocked flags;
the field interface encodes these flags rather than prerequisite-ID lists.\\
\end{longtable}
\endgroup

\section{Observation \& Perception}

\begin{figure*}[t]
  \centering
  \includegraphics[width=\linewidth]{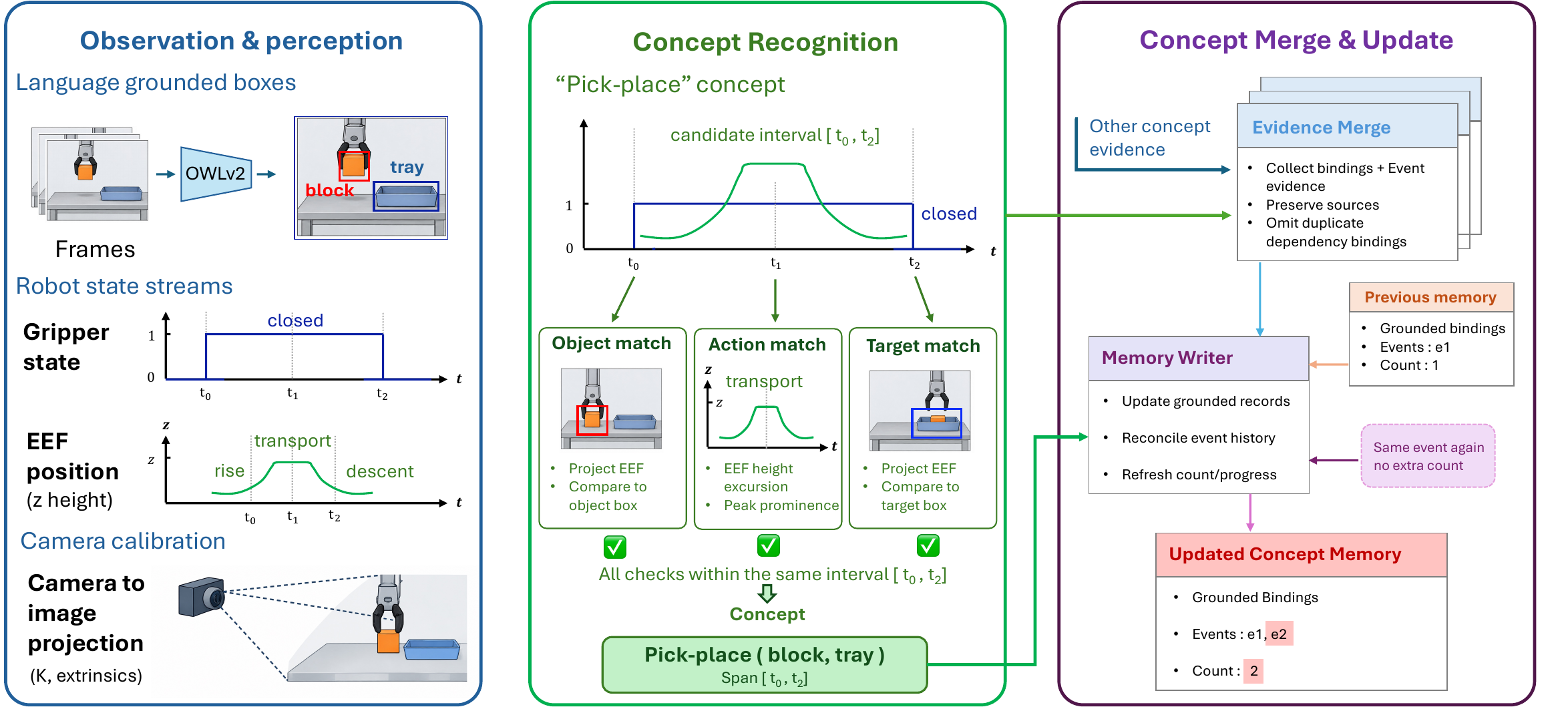}
  \caption{Concept recognition and memory update in ECoMEM. A pick-and-place interaction illustrates how language-grounded entities and robot-state evidence jointly establish an event. The writer merges concept evidence, updates grounded records, and advances progress once per distinct completion.}
  \label{fig:concept-recognition}
\end{figure*}

Figure~\ref{fig:concept-recognition} illustrates the recognition and
update process through a pick-and-place interaction. Its left and
middle panels show how shared perception outputs are composed into
concept evidence; the resulting memory update is described in
Appendix~\ref{app:memory-details}.

\subsection{Language-grounded perception.}
A shared, frozen OWLv2 detector~\cite{minderer2023owlv2} grounds
instruction-derived entity descriptions in camera observations.
Detections are associated with episode-local entity identities, and
their locations are normalized and quantized onto a shared spatial
grid.

\subsection{Temporal sampling and tracking.}
Recognition processes the available observation prefix. Keyframe
selection prioritizes gripper transitions and visual changes while
maintaining temporal coverage. Tracking uses image motion,
segmentation masks, and robot--object motion correspondence to update
entity associations. During occlusion, memory
retains the entity binding and its last-supported location.
Camera observations and robot states are aligned along the execution
timeline, including observations between policy queries. This aligned
stream supplies the evidence used by the selected concept recognizers.

\section{Concept Recognition}

\subsection{Concept selection.}
The concept library provides reusable recognition capabilities and
their dependencies. For all the tasks that have been seen, the model will distill a small and fast router to select concepts. This method was applied in the RoboMME simulation. However, for tasks that have not been seen before, or for combinations of different tasks, we have introduced a new option to utilize the LLM agent to select possible concepts. The task instructions
and library descriptions will guide an LLM to propose a Concept Plan:
a set of concepts to track during execution. Dependency expansion
adds the supporting concepts, and validation checks library membership,
dependency consistency, and required sensor inputs.
The instruction supplies entity descriptions, target counts, and
temporal constraints. These arguments instantiate the selected
recognizers. Selection specifies the concepts to track, and recognition
evaluates them against sensor evidence.

\subsection{Language-grounded concept recognition.}
The selected recognizers combine visual evidence with
end-effector motion, gripper state, and camera calibration to identify
relations, interaction events, and ordered bindings.
Each recognizer associates its result with grounded entities,
supporting evidence, and an observation time or interaction interval.
Spatial concepts describe entity configurations; event concepts
identify changes over an interval; temporal concepts organize
observed interactions.

\subsection{Composing evidence over an interaction.}
The middle panel of Figure~\ref{fig:concept-recognition} illustrates
three complementary checks. Gripper closure at $t_0$ and opening at
$t_2$ delimit a candidate interaction interval $[t_0,t_2]$.
\emph{Object match} projects the end-effector position at acquisition
into the image and compares it with the grounded object's box.
\emph{Action match} evaluates the candidate lift-and-transport motion
using the end-effector's vertical excursion and height peak above
the interval endpoints. \emph{Target match} compares the projected release
position with the grounded destination's box. Camera intrinsics and
extrinsics provide the common image-coordinate system for these checks.
Object and destination matching use detections available by their
respective interaction times.

When all checks hold for the same interval, the recognizer produces
a completed pick-and-place instance with its object, destination,
and temporal span. In the figure, these bindings are the block and
tray. Each concept specifies the entity, physical, and temporal
evidence used in its recognition rule.

\section{Concept Merge \& Update}
\label{app:memory-details}

The right panel of Figure~\ref{fig:concept-recognition} shows how
recognized concepts update a persistent memory. Evidence Merge
combines the selected recognizers' outputs; the Memory Writer
integrates them with previous records and refreshes event history
and progress. The Reader then converts these records into memory
tokens for the VLA policy.

\subsection{Evidence merging.}
Each recognizer produces entity bindings and supporting observations
with their sources and temporal references. Evidence Merge collects
these outputs and retains their provenance. Shared dependency evidence
is associated with its composite result, avoiding duplicate bindings
for intermediate stages. The merged evidence supplies the writer
with the selected concepts' spatial, relational, and temporal outputs.

\subsection{State estimation and commitment.}
For a grounded predicate identified by its predicate, subject,
optional object, and agent, the writer accumulates supporting and
opposing evidence over time. Let $L_t$ be its log-odds at update $t$,
initialized to zero. Channel $j$ contributes a signed score
$s_{j,t}\in[-1,1]$ with positive weight $w_j$.
Let $\Delta_t$ denote elapsed seconds since the previous update,
$h>0$ the decay half-life, $B>0$ the log-odds bound, and
$\delta_t=\min(\max(\Delta_t,10^{-9}),\delta_{\max})$ the
effective evidence duration, capped by $\delta_{\max}>0$.
The update is
\begin{equation}
L_t=\operatorname{clip}_{[-B,B]}
\left(2^{-\Delta_t/h}L_{t-1}
+\delta_t\sum_j w_j s_{j,t}\right),
\label{eq:appendix-filter}
\end{equation}
where the clipping operator bounds its argument to $[-B,B]$.
The first update uses $\Delta_t=0$.

Separate confirmation and retention thresholds provide hysteresis:
strong evidence establishes a state, while a lower level of support
maintains it. Opposing evidence can establish a false state, and
insufficient support returns the estimate to unknown. Attachment,
holding, and lifting additionally require physical evidence such as
tracked object motion or relative-pose consistency. Confirmed updates
replace the previous state for the same grounded predicate.
Mutually exclusive hypotheses, such as two objects held by one
gripper, are resolved using their accumulated support, with inferred
exclusions marked in the records.

\subsection{Event history and progress.}
The writer records newly confirmed predicate transitions and
recognized interaction completions. Event history is reconciled
across successive recognition updates using completion labels and
temporal correspondence. Evidence matched to an existing completion
updates that entry; a newly identified completion adds an event.
Unchanged state observations retain the existing
state; a later reversal and renewed transition can establish another
event instance.

In Figure~\ref{fig:concept-recognition}, the previous memory contains
one completed interaction, $e_1$. The newly recognized pick-and-place
interaction contributes a second event, $e_2$, increasing the count
from one to two. Evidence matched again to $e_2$ updates the existing
entry and leaves the count unchanged.
Completed units receive distinct bindings, so repetition progress
is represented by the collection of completed-unit records and their
operation or category associations.

Goal records specify desired grounded states and prerequisite goals.
Current-state goals are evaluated against the committed state;
persistent completion goals retain achieved progress. A dependent
completion requires matching evidence after its prerequisites have
been achieved. The writer updates satisfied, executable-frontier,
and blocked flags accordingly. These records expose estimated
world state, completed interactions, and goal progress to the policy.

\subsection{Record interface.}
Table~\ref{tab:record-fields} summarizes the eleven integer fields
supplied to the Reader. Entity slots are episode-local, and references
to the same entity share a slot across record types. Goals, beliefs,
events, groundings, and geometry records are serialized in that order,
retaining their within-type ordering. The memory store retains full
timestamps and evidence provenance; the Reader uses the corresponding
compact fields. Padding rows are excluded by a Boolean mask.

\begin{table}[htbp]
\centering
\small
\caption{Structured fields supplied to the memory Reader.
Subject/object and row/column each denote two fields.}
\label{tab:record-fields}
\setlength{\tabcolsep}{4pt}
\begin{tabular}{@{}p{0.22\linewidth}p{\dimexpr0.78\linewidth-2\tabcolsep\relax}@{}}
\hline
Field & Encoding\\
\hline
Kind & Record type: belief, event, goal, grounding, or geometry;
zero denotes padding.\\
Predicate & Shared identifiers for predicates, transition events,
entity roles, and categorical attributes.\\
Subject, object & Episode-local entity identifiers; zero denotes
an unspecified argument.\\
Truth & True, false, or unknown; zero denotes an inapplicable value.\\
Status & Tentative, confirmed, or stale; zero denotes an
inapplicable value.\\
Confidence & $\operatorname{round}(9p)$ for confidence $p\in[0,1]$.\\
Age & Zero for nonpositive age; otherwise
$\min(15,1+\lfloor\log_2 d\rfloor)$, where $d$ is the elapsed
number of frames since the record timestamp.\\
Flags & Satisfied, executable frontier, blocked, and derived
exclusion indicators.\\
Row, column & Cells 1--32 on a $32\times32$ image grid; zero
denotes an unspecified position.\\
\hline
\end{tabular}
\end{table}

Grounding records carry entity locations, and selected goal records
also carry their subject's target location. Relative ages and ordered
bindings encode temporal information, while completion records and
goal flags expose progress to the policy.

\subsection{Reader architecture.}
Each field is mapped to a 256-dimensional embedding. The eleven
embeddings are summed and layer-normalized to form one representation
per record. Two pre-normalized Transformer blocks integrate information
across valid records, using four attention heads and feed-forward
width 1,024 with GELU activations. Output normalization and a linear
projection produce one 2,048-dimensional token per record, scaled
by $\sqrt{2048}$. These memory tokens join the image and language
tokens in the VLA prefix. The Reader is trained jointly with the
policy, allowing action learning to use the maintained concept records.

\subsection{Causal training alignment.}
At each decision frame, the training sample pairs the current
observation and memory constructed through that frame with the
subsequent expert action chunk. Demonstrated references and execution
progress are maintained separately: a demonstration specifies what
to reproduce, while observed execution establishes what has been
completed. An event enters a sample's memory once its supporting
evidence is available. Entity-identifier augmentation applies a shared
bijection across all records in a sample, preserving the connections
between groundings, relations, events, and goals.

\section{Ablations of pipeline}

\subsection{Content interventions.}
\label{app:content-eval}
Figure~\ref{fig:content-ablations} modifies the record interface, with
current images, robot inputs, action targets, and task requirements
held fixed within each pair.
\begin{itemize}
\item \textbf{No Spatial (VideoUnmask):} set row and column fields to
zero in every active record, including any goal or geometry row
carrying coordinates. All other fields and masks are unchanged.
\item \textbf{No Temporal (PatternLock):} preserve the active fact
multiset and spatial cells, zero every age, consistently remap entity
slots by grounding location, and lexicographically sort the resulting
rows before packing the mask. Grounding ties use predicate and then
original-slot tie-breaks. Progress flags and fact multiplicities remain.
This intervention removes relative-age and ordered-binding cues from
the record presentation.
\item \textbf{No Event/Progress (PickXtimes):} remove event rows and
beliefs encoding cycle or procedure completion, clear the satisfied,
frontier, and blocked bits, remove groundings referenced only by
deleted records, zero ages, canonicalize entity slots, and sort/pack
the surviving rows. The positive condition exposes semantic cycle
and procedure completion events while excluding low-level contact,
holding, and transport event rows from the Reader. Both conditions
retain the current-state evidence needed by recognition.
\end{itemize}

For an operational example, a grounding row
\texttt{(4, role:object, 2, 0, 0, 0, 9, 0, 0, 12, 21)}
becomes
\texttt{(4, role:object, 2, 0, 0, 0, 9, 0, 0, 0, 0)}
under No Spatial; the predicate name denotes its integer code.
Under No Temporal, nonzero belief/event ages become 0, entity slots
are consistently renamed, and records are re-sorted. Under No
Event/Progress, completion events are removed, surviving records are
packed contiguously, and goal-progress bits are cleared.

\subsection{Initialization latency.}
Validation episodes 0--4 are timed for each condition, sequentially
on the same pair of RTX 3090 GPUs. The timer starts before the
recognition call and ends after evidence merging, memory construction,
and interface validation. The measured interval includes first-use
detector loading and warmup occurring within these boundaries.
VLA loading and action execution occur outside the timed interval.
The median of five episodes is 2.23\,s versus 40.47\,s for
VideoUnmask and 4.38\,s versus 52.66\,s for VideoRepick.
Corresponding means are 3.87/41.41\,s and 5.51/56.61\,s.
These measurements refer to per-episode initialization.

Table~\ref{tab:initialization-latency} collects these measurements.
The ratios in the main text compare median per-episode initialization
times: No Selection divided by task-conditioned selection.

\begin{table}[htbp]
\centering
\small
\caption{Memory initialization on matched episodes using the same
pair of RTX 3090 GPUs. Times are in seconds; each condition uses
five episodes. C denotes task-conditioned selection and A denotes
No Selection.}
\label{tab:initialization-latency}
\begin{tabular}{lrrrrr}
\hline
Task & Median C & Median A & A/C & Mean C & Mean A\\
\hline
VideoUnmask &2.23&40.47&$18.1\times$&3.87&41.41\\
VideoRepick &4.38&52.66&$12.0\times$&5.51&56.61\\
\hline
\end{tabular}
\end{table}

\subsection{Language-conditioned grounding.}
\label{app:grounding-eval}
The No Language intervention replaces entity phrases with role-level
queries: covers, containers, locations, and regions use
\emph{a container}; controls, buttons, and waypoints use \emph{a control};
pegs use \emph{a peg}; remaining entities use \emph{an object}.
For example, \emph{a green cube} becomes \emph{an object}, while a
PatternLock request containing \emph{a stick}, \emph{a button},
\emph{a pad}, \emph{a marker}, and \emph{a circle} becomes
\emph{an object} followed by four \emph{a control} queries.
Concept selection, required counts, role assignments, detector,
keyframe budget, and policy instruction remain fixed.
Policy evaluation includes episodes with empty or partial memory
resulting from unavailable groundings.

Offline evaluation measures agreement with pipeline-generated
reference records extracted from stored full-memory outputs before
the query intervention.
Target-location agreement compares exact $8\times8$ cells: a
nonzero Reader coordinate $c$ is mapped to
$1+\lfloor(c-1)/4\rfloor$. Sequence agreement requires the same
length and exact cell agreement at every sequence position.
VideoUnmask uses the first 100 archived episodes, with 98/100 matches
for specific queries and 20/100 for generic queries.
PatternLock uses 98: episodes 29 and 83 lack the required reference
row at their retrieval frame and are excluded from both conditions.
Matches are 86/98 and 0/98.
Separately, 50 matched closed-loop episodes give 43/50 versus 12/50
for VideoUnmask and 36/50 versus 1/50 for PatternLock.

Table~\ref{tab:grounding-counts} reports both the reference-matching
counts and the separate closed-loop evaluation counts.

\begin{table}[htbp]
\centering
\small
\caption{Language-grounding results. C uses specific entity queries;
A uses generic role-level queries. Matches denote agreement with
pipeline-generated reference records for the target cell (VideoUnmask)
or complete ordered sequence (PatternLock). Successes denote
closed-loop task completion.}
\label{tab:grounding-counts}
\begin{tabular}{lrrrr}
\hline
Task & Matches C & Matches A & Successes C & Successes A\\
\hline
VideoUnmask &98/100&20/100&43/50&12/50\\
PatternLock &86/98&0/98&36/50&1/50\\
\hline
\end{tabular}
\end{table}

\section{VLM-based Concept Extraction}
\label{app:vlm_concept_extraction}

\subsection{Dataset and supervision.}
We construct a dataset of 504 recorded RoboMME videos from six tasks:
VideoRepick, PickHighlight, PickXtimes, StopCube,
VideoPlaceOrder, and VideoPlaceButton.
The dataset covers eight concepts: picked-object identity,
highlighted-object sets, current holding and highlighted states,
pick--place counts, region-entry counts, placement order,
and placement relations around button presses.
We use 365 training, 44 validation, and 95 test videos,
keeping scene-seed groups within a single split.
Reference labels are generated from independent simulator replays
using observed physical events, rendered highlights, and button states.
Replay alignment is verified against recorded RGB, depth,
robot poses, and camera calibration.
Labels are algorithmically generated and audited,
rather than exhaustively hand-annotated.

Historical concepts use complete video clips, whereas current-state
concepts query a named object at a video-prefix endpoint.
For state supervision, we select one stable positive and one stable
negative endpoint per eligible video, producing 808 queries from
404 of the same videos: 590 training, 68 validation, and 150 test queries.
Holding labels require bilateral finger contact and object lift;
highlighted-state labels require an active visible ring.
Both states must persist over five frames.
All derived queries retain their source video's split.
Each input contains a concept definition and 64 uniformly sampled
front-camera RGB frames at $256\times256$ resolution,
restricted to the queried prefix for current-state questions.
Objects are referenced through first-frame appearance or spatial
descriptions; reference boxes, instance IDs, and simulator states
are not provided to the model.

\subsection{Training.}
We compare two matched formulations:
\emph{Structured} generates JSON records, and \emph{QA} generates
concept-specific textual answers, including yes/no for state queries.
Both use identical visual inputs, semantic supervision, and sample
order within each comparison.
We train separate adapters for historical and current-state concepts
under each formulation, all initialized from the original
Qwen3-VL-8B-Instruct checkpoint.
Only language-model attention LoRA parameters are updated
($q,k,v,o$ projections; rank~8, scaling~16, dropout~0);
all remaining parameters are frozen.
Training uses BF16, answer-token cross-entropy, AdamW with a constant
learning rate of $10^{-4}$, weight decay $0.01$, and gradient clipping
at 1.0.
Historical-concept training comprises 915 updates and 1,830 video
presentations per formulation, including one distributed-sampling
padding example per epoch.
Current-state training comprises 1,475 updates and 2,950 query
presentations per formulation.
Each checkpoint is selected by the lowest validation answer-token
negative log-likelihood, without test-based selection.

\section{Real-Robot Experimental Details}
\label{app:real-robot}

\subsection{Tasks and data.}
We evaluate \textit{ScoopPour} and \textit{CupSwitch} using the right
arm of a YAM robot. ScoopPour requires scooping red beans from a tray
with a yellow scoop and pouring them onto a plate. CupSwitch requires
moving a purple cup between two plates and back; one round trip counts
as one cycle. Demonstrations record top-view and right-wrist RGB images
($640\times480$), right-arm state, and control commands at a nominal
30\,Hz. State and action each contain six joint coordinates and one
gripper coordinate.

After screening, each task uses 59 demonstrations: 80,203 frames for
ScoopPour and 81,200 for CupSwitch. The respective two-/three-cycle
splits are 24/35 and 29/30 demonstrations. Both methods use the same
retained data, with English instructions specifying the task and count.
All retained demonstrations are used for fine-tuning. Evaluation covers
ScoopPour $1$--$3\times$ and CupSwitch $2$--$3\times$; the one-cycle
ScoopPour target is absent from the fine-tuning data.

\subsection{Concept adaptation and memory generation.}
\label{app:enrichment}
\label{app:enrichment-eval}
For ScoopPour, we introduce \texttt{event.scoop} while reusing entity grounding, event ordering, and counting.
OWLv2 grounds the scoop, tray, and plate; top-view tracking estimates
location, and wrist-view appearance indicates whether the scoop contains
beans. A loaded scoop after visiting the tray establishes an acquisition
candidate. A subsequent loaded-to-empty transition above the plate,
confirmed across consecutive observations, completes one cycle. CupSwitch reuses entity, identity, pick-and-place,
sequence, and counting concepts to compose two directed transfers into
a round trip.

Agent-assisted visual review and user feedback guide development on
six ScoopPour and two CupSwitch demonstrations. We inspect six additional
demonstrations per task; the ScoopPour recognizer is frozen before this
check. Review compares visible states, event evidence, cycle counts,
and confirmation timing. Logic tests and evidence-removal replays check
event order, repeated counting, and required observations.

The recognizers automatically generate causal, frame-aligned memory for
all retained demonstrations. The existing 11-field interface encodes
entity locations, current states and relations, observed events, and
goal progress. Current states update with observations, while completed
events retain their confirmation frames. Each policy input includes only
memory available by that frame; the instruction supplies the requested
count, and recognized events supply observed progress.

\subsection{Policy training.}
We train four task-specific policies: Concept Memory and a no-memory
$\pi_{0.5}$ baseline for each task. Shared backbones initialize from
official pretrained $\pi_{0.5}$ weights and undergo full-parameter
fine-tuning.
Both methods use identical task data, inputs, and normalization;
the baseline omits the memory encoder and tokens.
Images are letterboxed to $224\times224$, and state/actions use
first- and 99th-percentile normalization. Dense-frame sampling supplies
20-step action targets, masking unused dimensions and episode-tail padding.

\subsection{Deployment and evaluation.}
A remote GPU server predicts 20-step action chunks, and the local robot
executes the first 16 steps before replanning. Memory updates from streamed observations at
configured rates of 5\,Hz for ScoopPour and 10\,Hz for CupSwitch and
resets at episode boundaries. Both methods use the same execution
interface and command-increment limits.

\subsection{Results.}
Table~\ref{tab:detailed_real_robot_success_rate} reports success rates
across the five conditions. Concept Memory achieves 68/79 successes
(86.08\%), versus 5/58 (8.62\%) for the baseline, a gain of 77.46
percentage points. Their pooled Wilson 95\% confidence intervals are
$[76.76,92.04]\%$ and $[3.74,18.64]\%$, respectively
(two-sided Fisher exact test: $p=4.90\times10^{-21}$).
Table~\ref{tab:failure_breakdown} details the manual outcome annotations.

\begin{table*}[t]
\centering
\small
\caption{Real-robot success rates by task and target count.
Targets count scoop-and-pour cycles for ScoopPour and complete round trips
for CupSwitch. Gain is measured in percentage points (pp).}
\label{tab:detailed_real_robot_success_rate}
\setlength{\tabcolsep}{4pt}
\renewcommand{\arraystretch}{1.12}
\begin{tabular*}{\linewidth}{@{\extracolsep{\fill}}lcrrrrr@{}}
\toprule
& & \multicolumn{2}{c}{Concept Memory}
& \multicolumn{2}{c}{No-memory $\pi_{0.5}$} & \\
\cmidrule(lr){3-4}\cmidrule(lr){5-6}
Task & Target & Success / Trials & Rate (\%)
& Success / Trials & Rate (\%) & Gain (pp) \\
\midrule
ScoopPour & $1\times$ & 16 / 20 & 80.00 & 4 / 12 & 33.33 & +46.67 \\
& $2\times$ & 19 / 23 & 82.61 & 0 / 12 & 0.00 & +82.61 \\
& $3\times$ & 11 / 12 & 91.67 & 0 / 11 & 0.00 & +91.67 \\
\addlinespace[4pt]
CupSwitch & $2\times$ & 11 / 12 & 91.67 & 0 / 12 & 0.00 & +91.67 \\
& $3\times$ & 11 / 12 & 91.67 & 1 / 11 & 9.09 & +82.58 \\
\midrule
\multicolumn{7}{@{}l}{\textit{Aggregates over reported conditions}} \\
\addlinespace[2pt]
ScoopPour & $1$--$3\times$
& 46 / 55 & 83.64 & 4 / 35 & 11.43 & +72.21 \\
CupSwitch & $2$--$3\times$
& 22 / 24 & 91.67 & 1 / 23 & 4.35 & +87.32 \\
\addlinespace[3pt]
\multicolumn{2}{@{}l}{\textbf{Pooled success rate}}
& \textbf{68 / 79} & \textbf{86.08}
& \textbf{5 / 58} & \textbf{8.62} & \textbf{+77.46} \\
\multicolumn{2}{@{}l}{Mean over five conditions}
& -- & 87.52 & -- & 8.48 & +79.04 \\
\bottomrule
\end{tabular*}

\vspace{4pt}
\begin{minipage}{\linewidth}
\footnotesize
Reported conditions comprise ScoopPour $1$--$3\times$ and CupSwitch $2$--$3\times$.
The mean weights these five conditions equally; the pooled rate weights
individual trials equally.
\end{minipage}
\end{table*}

\begin{table*}[t]
\centering
\small
\caption{Manual real-robot outcome annotations. Numeric entries give
the number of trials. Baseline outcomes include successes and failures.}
\label{tab:failure_breakdown}
\setlength{\tabcolsep}{5pt}
\renewcommand{\arraystretch}{1.16}
\begin{tabularx}{\linewidth}{@{}lcl>{\raggedright\arraybackslash}X>{\raggedright\arraybackslash}X@{}}
\toprule
& & & \multicolumn{2}{c}{No-memory $\pi_{0.5}$ outcomes} \\
\cmidrule(lr){4-5}
Task & Target & \shortstack[l]{Concept Memory\\failures}
& Recorded cycle counts & Other outcomes \\
\midrule
ScoopPour & $1\times$ & 4 failures
& 1 (success): 4;\newline 2: 2; 6: 1
& Too many cycles: 2;\newline timeout: 1; freeze: 1; miss: 1 \\
\addlinespace[4pt]
& $2\times$ & 4 failures
& 0: 2; 1: 6;\newline 3: 1; 4: 1
& Timeout: 2 \\
\addlinespace[4pt]
& $3\times$ & 1 failure
& 2: 2; 4: 2
& Timeout: 7 \\
\midrule
CupSwitch & $2\times$ & 1 timeout
& 0: 6;\newline prolonged 0-cycle execution: 1
& Timeout: 4; miss: 1 \\
\addlinespace[4pt]
& $3\times$ & 1 timeout
& 0: 1; 1: 2
& Too many cycles: 2;\newline timeout: 3; miss: 1;\newline
3 cycles then timeout: 1;\newline success: 1 \\
\bottomrule
\end{tabularx}

\vspace{4pt}
\begin{minipage}{\linewidth}
\footnotesize
\textit{Notation:} An entry of \emph{2: 4} means four trials annotated
with two completed cycles. \emph{Too many cycles} retains annotations
without an exact count.
\textit{Scoring:} Concept Memory trials marked \texttt{fail} or
\texttt{timeout} are failures; other annotations indicate success.
For the baseline, a numeric suffix matching the requested count or
an absent suffix indicates success. Textual failure annotations,
including timeouts after reaching the target count, indicate failure.
\end{minipage}
\end{table*}

\end{document}